\documentclass[10pt,twocolumn,letterpaper]{article}

\usepackage{cvpr}              

\usepackage[ruled]{algorithm2e}
\usepackage[accsupp]{axessibility} 
\usepackage{colortbl}
\usepackage[symbol]{footmisc}
\usepackage{multirow}
\usepackage{pifont}
\usepackage[dvipsnames]{xcolor}

\newcommand{\eref}[1]{Eq.~\eqref{#1}}
\newcommand{\Eref}[1]{Equation~\eqref{#1}}
\newcommand{\fref}[1]{Fig.~\ref{#1}}
\newcommand{\Fref}[1]{Figure~\ref{#1}}
\newcommand{\tref}[1]{Tab.~\ref{#1}}
\newcommand{\Tref}[1]{Table~\ref{#1}}
\newcommand{\best}[1]{\textbf{#1}}
\newcommand{\second}[1]{\underline{#1}}
\renewcommand{\paragraph}[1]{\vspace{2.5pt}\noindent\textbf{#1}}

\definecolor{color1}{HTML}{FF1F5B}
\definecolor{color2}{HTML}{009ADE}
\definecolor{color3}{HTML}{FFC61E}

\definecolor{cvprblue}{rgb}{0.21,0.49,0.74}
\usepackage[pagebackref,breaklinks,colorlinks,citecolor=cvprblue]{hyperref}

\def\paperID{*****} 
\def\confName{CVPR}
\def\confYear{2024}

\title{Unified Entropy Optimization for Open-Set Test-Time Adaptation}

\author{Zhengqing Gao$^{1,2}$\quad Xu-Yao Zhang$^{1,2}$\footnotemark[1]{}\quad Cheng-Lin Liu$^{1,2}$\\
$^1$MAIS, Institute of Automation, Chinese Academy of Sciences\\
$^2$School of Artificial Intelligence, University of Chinese Academy of Sciences\\
{\tt\small gaozhengqing2021@ia.ac.cn\quad\{xyz, liucl\}@nlpr.ia.ac.cn}
}

\begin{document}
\maketitle
\footnotetext[1]{Corresponding author.}
\begin{abstract}
Test-time adaptation (TTA) aims at adapting a model pre-trained on the labeled source domain to the unlabeled target domain. Existing methods usually focus on improving TTA performance under covariate shifts, while neglecting semantic shifts. In this paper, we delve into a realistic open-set TTA setting where the target domain may contain samples from unknown classes. Many state-of-the-art closed-set TTA methods perform poorly when applied to open-set scenarios, which can be attributed to the inaccurate estimation of data distribution and model confidence. To address these issues, we propose a simple but effective framework called unified entropy optimization (UniEnt), which is capable of simultaneously adapting to covariate-shifted in-distribution (csID) data and detecting covariate-shifted out-of-distribution (csOOD) data. Specifically, UniEnt first mines pseudo-csID and pseudo-csOOD samples from test data, followed by entropy minimization on the pseudo-csID data and entropy maximization on the pseudo-csOOD data. Furthermore, we introduce UniEnt+ to alleviate the noise caused by hard data partition leveraging sample-level confidence. Extensive experiments on CIFAR benchmarks and Tiny-ImageNet-C show the superiority of our framework. The code is available at \url{https://github.com/gaozhengqing/UniEnt}.
\end{abstract}    
\section{Introduction}
\label{sec:intro}

\begin{figure}[t]
    \centering
    \includegraphics[width=\linewidth]{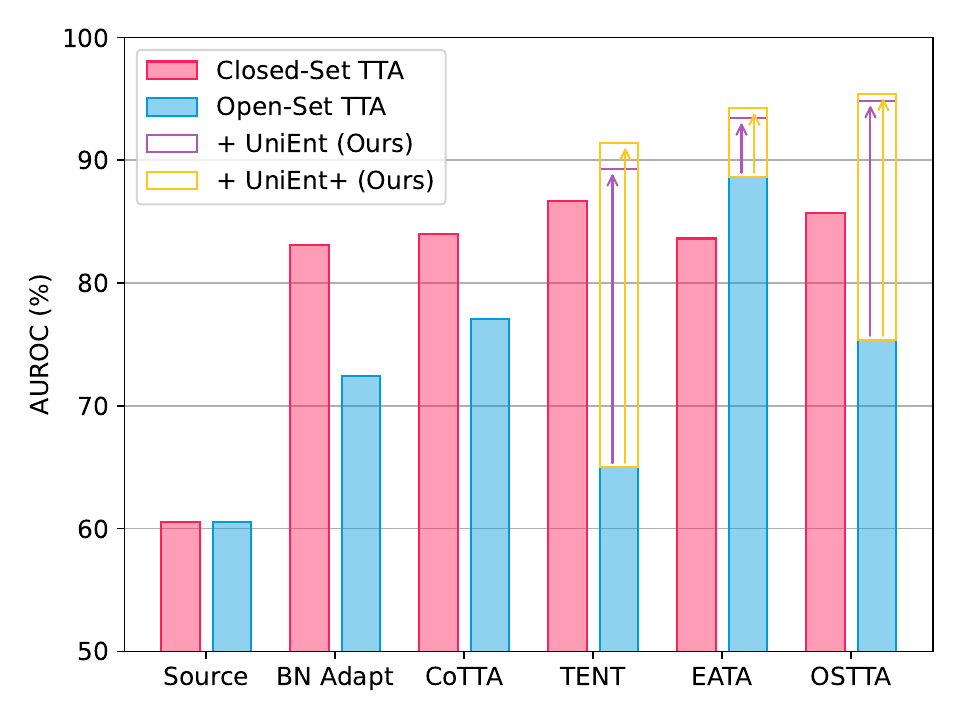}
    \caption{Existing TTA methods exhibit performance degradation with unknown classes included, while our methods can improve them significantly. We compare BN Adapt~\cite{nado2020evaluating}, CoTTA~\cite{wang2022continual}, TENT~\cite{wang2021tent}, EATA~\cite{niu2022efficient}, and OSTTA~\cite{lee2023towards}.}
    \label{fig:auroc}
\end{figure}

Deep neural networks (DNNs) have achieved great success in recent years when the training and test data are drawn \iid from the same distribution. However, in many real-world applications, this strict assumption is difficult to hold. Models deployed in practice can encounter different types of distribution shifts. On the one hand, the model needs to be able to address semantic shifts, \ie, identify samples from unknown classes, which has given rise to problems such as out-of-distribution (OOD) detection~\cite{hendrycks2017baseline,hendrycks2019deep,liu2020energy,hendrycks2022scaling,zhang2023openood} and open-set recognition~\cite{chen2020learning,chen2021adversarial,vaze2022open,huang2022class}. On the other hand, the model needs to be robust to covariate shifts and have good generalization performance to different styles and domains. Many efforts have been devoted to reduce the performance gap of DNNs under covariate shifts, such as domain generalization~\cite{zhou2021domain,xu2021fourier,zhou2022domain,wang2022generalizing} and domain adaptation~\cite{ganin2016domain,yang2020fda}. Among various studies addressing covariate shifts, test-time adaptation (TTA) has recently received increasing attention because its practicality: neither source domain data nor target domain labels are required~\cite{nado2020evaluating,wang2021tent,wang2022continual,niu2022efficient,lee2023towards,li2023robustness}.

\begin{figure}[t]
    \centering
    \includegraphics[width=\linewidth]{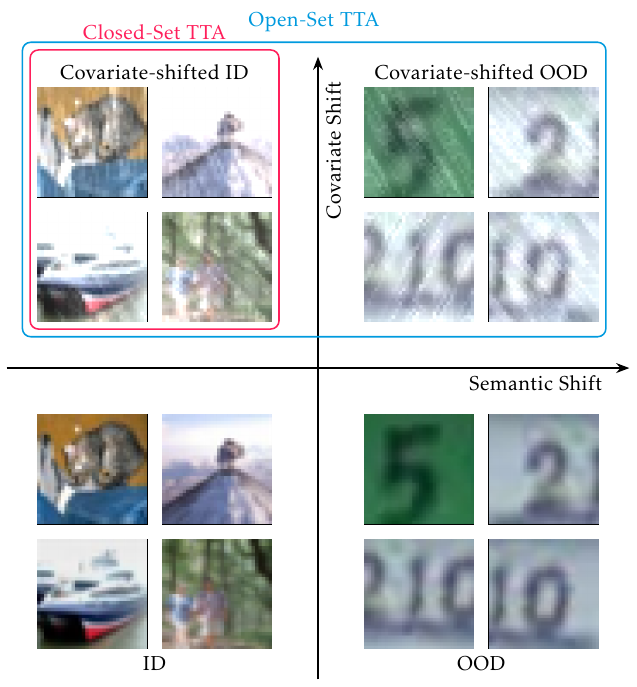}
    \caption{Comparison between closed-set TTA and open-set TTA.}
    \label{fig:open_set_tta}
\end{figure}

Nevertheless, most of the existing TTA methods~\cite{nado2020evaluating,wang2021tent,wang2022continual,niu2022efficient} focus only on solving the covariate shift and ignoring the semantic shift. We believe that this is impractical since we cannot guarantee the test samples contain only the classes seen in the training phase. Many recent works~\cite{lee2023towards,li2023robustness} have realized this and made some initial attempts. \Fref{fig:open_set_tta} illustrates the differences between the traditional closed-set TTA and the novel open-set TTA settings. First, we need to clarify that in the literature on OOD detection, out refers specifically to ``outside the semantic space", whereas in the literature on OOD generalization, out refers specifically to ``outside the covariate space". Here we follow the terminology used in~\cite{zhang2023openood}. According to the different types of distribution shifts, we divide the real-world data into four types:
\begin{itemize}
    \item In-distribution (ID) data is the most common data we typically use to train a model, with a limited number of classes.
    \item Out-of-distribution (OOD) data contains some open classes that have not been seen before in ID data, with the same style and domain as ID data.
    \item Covariate-shifted ID (csID) data and ID data have the same classes and differ in styles and domains.
    \item Covariate-shifted OOD (csOOD) data is different from ID data in both classes and domains.
\end{itemize}
The open-set TTA setting takes into account both csID data and csOOD data.

Existing TTA methods make extensive use of entropy objective, which proves to be very effective. We first experimentally verify that existing TTA methods~\cite{zagoruyko2016wide,nado2020evaluating,wang2021tent,wang2022continual,niu2022efficient,lee2023towards} degrade the classification accuracy of known classes when open-set classes are included, which is consistent with the conclusions drawn from some recent studies~\cite{lee2023towards,li2023robustness}. In addition, as shown in \fref{fig:auroc}, the detection performance of unknown classes is also impaired, which has not received enough attention in previous studies. We attribute the performance degradation to the following two points. First, the presence of open-set samples leads to the incorrect estimation of normalization statistics by the model, leading to errors in updating affine parameters. Second, entropy minimization on samples from unknown classes forces the model to output confident predictions, undermining the model's confidence and leading to a decrease in the model's ability to distinguish between known classes and unknown classes.

With the aforementioned causes in mind, we propose three techniques to enhance the robustness of existing TTA methods under open-set setting. We first propose a distribution-aware filter to preliminarily distinguish between csID samples and csOOD samples. Specifically, we observe that the cosine similarity between the features extracted by the source model and the source domain prototypes can reflect the semantic shift, and we use this property to distinguish samples. We then propose a unified entropy optimization framework (UniEnt) to address the aforementioned challenges. UniEnt minimizes the entropy of csID samples while maximizing the entropy of csOOD samples simultaneously. Furthermore, we propose UniEnt+ using a sample-level weighting strategy to avoid the error caused by noisy data partition.

We summarize the contributions of this paper as follows.
\begin{itemize}
    \item We first delve into the performance of existing methods under closed-set TTA and open-set TTA settings. We then summarize two reasons for the performance degradation of existing methods with open-set classes included.
    \item We propose a unified entropy optimization framework, which consists of a distribution-aware filter to distinguish csID and csOOD samples, entropy minimization on csID samples to obtain good classification performance on known classes and entropy maximization on csOOD samples to obtain good detection performance on unknown classes.
    \item Our proposed framework can be flexibly applied to many existing TTA methods and substantially improves their performance under open-set setting. Comprehensive experiments demonstrate the effectiveness of our approach.
\end{itemize}

\begin{figure*}[t]
    \centering
    \includegraphics[width=\linewidth]{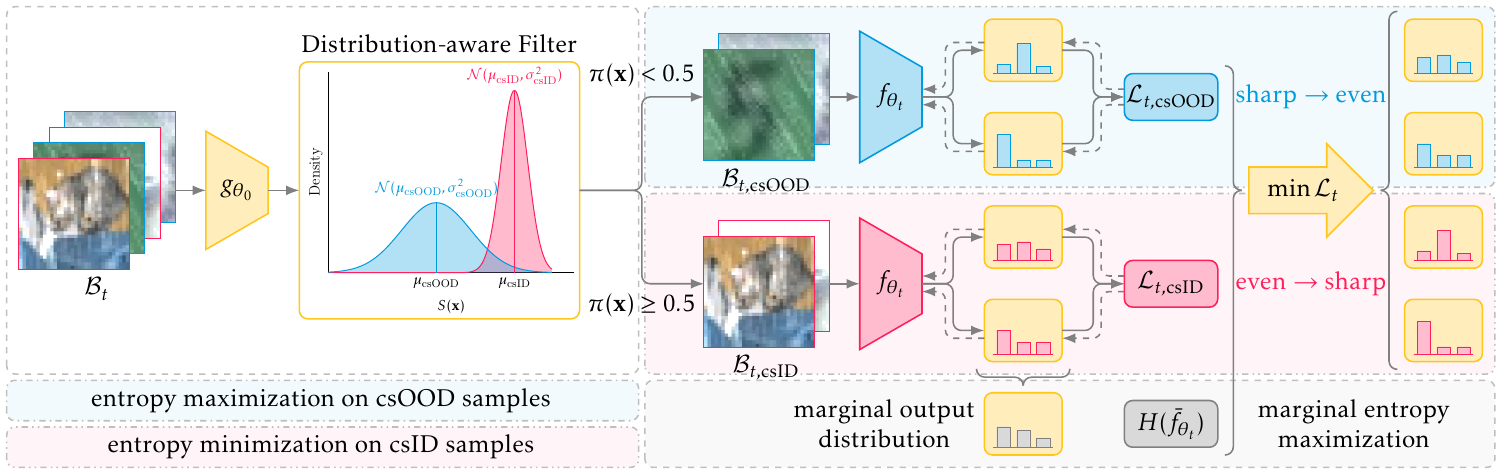}
    \caption{Illustration of the unified entropy optimization (UniEnt) framework. At timestamp $t$, mini-batch $\mathcal{B}_t$ may contain samples from csID and csOOD. First, we filter csOOD samples by csOOD score $S(\mathbf{x})$. Then, we perform entropy minimization for csID samples and entropy maximization for csOOD samples, we also adopt marginal entropy maximization to pervent model collapse. After optimization, we can yield better classification and detection performance tradeoff.}
    \label{fig:framework}
\end{figure*}

\section{Related Work}
\label{sec:relatedwork}

\paragraph{Test-time adaptation.} Among all the approaches to solving covariate shifts, test-time adaptation has received much attention because of its challenging setting of accessing only the source model and unlabelled target data. Some of the initial work~\cite{nado2020evaluating,schneider2020improving,wang2021tent,you2021test,khurana2021sita,lim2023ttn} focused on improving TTA performance by estimating batch normalization statistics using test data and designing unsupervised objective functions, \eg, TENT~\cite{wang2021tent} proposed to optimize the affine parameters of batch normalization by minimizing the entropy of model outputs. These works mainly focus on static TTA and do not take into account the changes in the domain. After adapting to a target domain, the adapted model is reset to the one pretrained on the source domain to adapt to the next domain. Later, some work~\cite{wang2022continual,niu2022efficient} proposed the continual TTA setting where the model needs to adapt to a series of continuously changing target domains without knowing the domain labels. This poses new challenges for TTA: catastrophic forgetting and error accumulation. CoTTA~\cite{wang2022continual} addresses the above issues through teacher-student model structure with data augmentation and stochastic recovery, while EATA~\cite{niu2022efficient} addresses the above issues through sample selection and anti-forgetting regularizer.

\paragraph{Robust test-time adaptation.} Recently, several works have paid more attention to the robustness of TTA methods. LAME~\cite{boudiaf2022parameter}, NOTE~\cite{gong2022note} and RoTTA~\cite{yuan2023robust} focus on the performance of TTA methods under non-\iid correlated sampling of test data. SITA~\cite{khurana2021sita} and MEMO~\cite{zhang2022memo} explore techniques for performing TTA on a single image. ODS~\cite{zhou2023ods} addresses case with label shift. OSTTA~\cite{lee2023towards} pays attention to the performance degradation caused by long-term TTA. OWTTT~\cite{li2023robustness} and OSTTA~\cite{lee2023towards} consider the scenarios where the test data includes unknown classes. SAR~\cite{niu2023towards} comprehensively analyzed the impact of mixed domain shifts, small batch sizes, and online imbalanced label distribution shifts on TTA performance. It is worth noting that there are some differences between the settings proposed by OWTTT~\cite{li2023robustness} and OSTTA~\cite{lee2023towards}, the samples of unknown classes in OWTTT~\cite{li2023robustness} are drawn from OOD, while the samples of unknown classes in OSTTA~\cite{lee2023towards} are drawn from csOOD. We adopt the setting proposed in OSTTA~\cite{lee2023towards} because of its practicality and challenging nature. First, the unknown class samples we encounter during TTA are likely to experience the same covariate shift. Second, it is more difficult to distinguish between csID samples and csOOD samples than between csID samples and OOD samples.

\paragraph{OOD detection.} For models deployed in real-world scenarios, the ability of OOD detection is crucial. Recent studies in OOD detection can be roughly divided into two categories. The first type of approaches~\cite{hendrycks2017baseline,liang2018enhancing,hsu2020generalized,liu2020energy,hendrycks2022scaling} is devoted to design sophisticated score functions and input-output transformations. MSP~\cite{hendrycks2017baseline} uses the maximum softmax probability to detect OOD samples. ODIN~\cite{liang2018enhancing} and generalized ODIN~\cite{hsu2020generalized} further introduces temperature scaling, input preprocessing and confidence decompose to improve OOD detection performance. The second type of approaches instead regularizes the model by exploring the additional outlier data~\cite{hendrycks2019deep,yu2019unsupervised,yang2021semantically,katz2022training,zhang2023mixture}. For example, OE~\cite{hendrycks2019deep} encourages the model to output low-confidence predictions for anomalous data. WOODS~\cite{katz2022training} on the other hand utilizes unlabelled wild data to improve the detection performance. SCONE~\cite{bai2023feed} considers both OOD detection and OOD generalization for the first time. It is worth noting that all the methods mentioned above are designed for the training phase. Recently, AUTO~\cite{yang2023auto} propose to optimize the network using unlabeled test data at test time to imporve OOD detection performance.

\section{Methodology}
\label{sec:methodology}

\subsection{Problem Setup}

Let $\mathcal{D}_s=\{\mathbf{x}_i,y_i\}_{i=1}^{N_s}$ be the source domain dataset with label space $\mathcal{Y}_s=\{1,\cdots,C_s\}$, and $\mathcal{D}_t=\{\mathbf{x}_j,y_j\}_{j=1}^{N_t}$ be the target domain dataset with label space $\mathcal{Y}_t=\{1,\cdots,C_t\}$, where $C_s$ and $C_t$ denote the number of classes in the source and target domain datasets, respectively. $C_s$ is equal to $C_t$ for closed-set TTA while $C_s<C_t$ always holds for open-set TTA. Given a model $f_{\theta_0}$ pre-trained on $\mathcal{D}_s$, TTA aims to adapt the model to $\mathcal{D}_t$ without target labels accessible. To be specific, we denote the mini-batch of test samples at timestamp $t$ as $\mathcal{B}_t$ and the adapted model as $f_{\theta_t}$. The main objective of open-set TTA is to correctly predict the classes in $\mathcal{Y}_s$ while reject the classes in $\mathcal{Y}_t\setminus\mathcal{Y}_s$ using the adapted model $f_{\theta_t}$, especially in the presence of large data distribution shifts.

\subsection{Preliminaries}

For closed-set TTA, a common practice~\cite{wang2021tent} is to adapt the model by minimizing the unsupervised entropy objective:
\begin{equation}
    \min_{\theta_t}\mathcal{L}_t=\frac{1}{\|\mathcal{B}_t\|}\sum_{\mathbf{x}\in\mathcal{B}_t}H(f_{\theta_t}(\mathbf{x}))-\lambda H(\bar{f}_{\theta_t}),
\end{equation}
where $H(f_{\theta_t}(\mathbf{x}))=-\sum_{c=1}^{C}f_{\theta_t}^c(\mathbf{x})\log f_{\theta_t}^c(\mathbf{x})$ denotes the entropy of the softmax output $f_{\theta_t}(\mathbf{\mathbf{x}})$, $\bar{f}_{\theta_t}=\frac{1}{\|\mathcal{B}_t\|}\sum_{\mathbf{x}\in \mathcal{B}_t}f_{\theta_t}(\mathbf{x})$ represents the average softmax output over the mini-batch $\mathcal{B}_t$, and $\lambda$ is a hyperparameter used to balance the two terms in the loss function. In previous studies~\cite{liang2020we,chen2022contrastive,choi2022improving,lim2023ttn}, marginal entropy $H(\bar{f}_{\theta_t})$ has been widely adopted to prevent model collapse, \ie, predicting all input samples to the same class.

\subsection{Motivation}

\begin{figure}[t]
    \centering
    \subfloat[Acc$\uparrow$]{\includegraphics[width=0.48\linewidth]{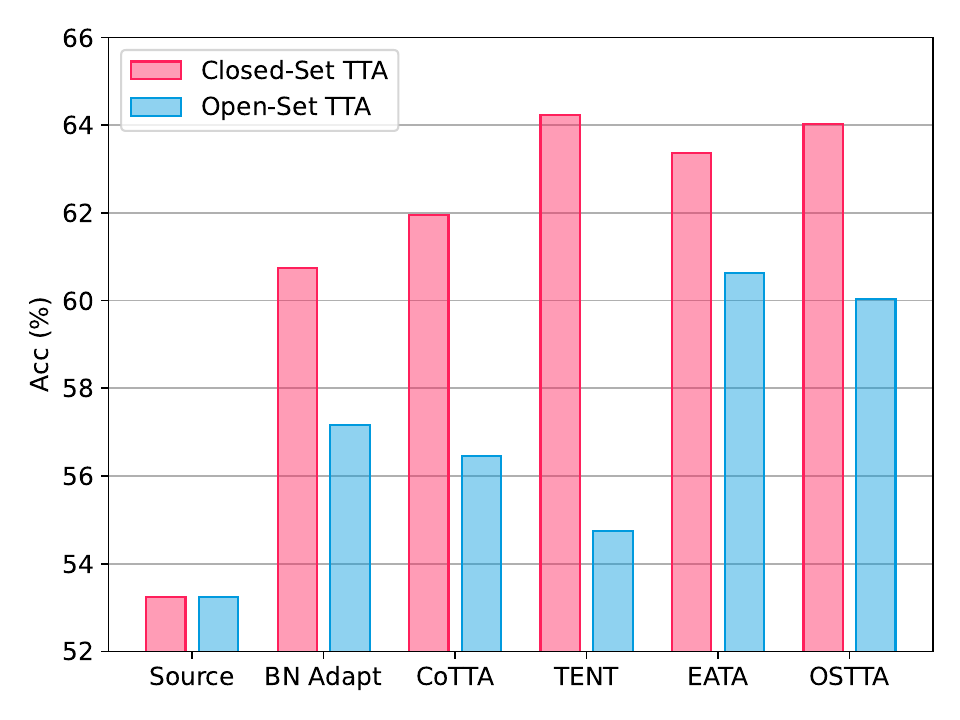}}
    \subfloat[FPR@TPR95$\downarrow$]{\includegraphics[width=0.48\linewidth]{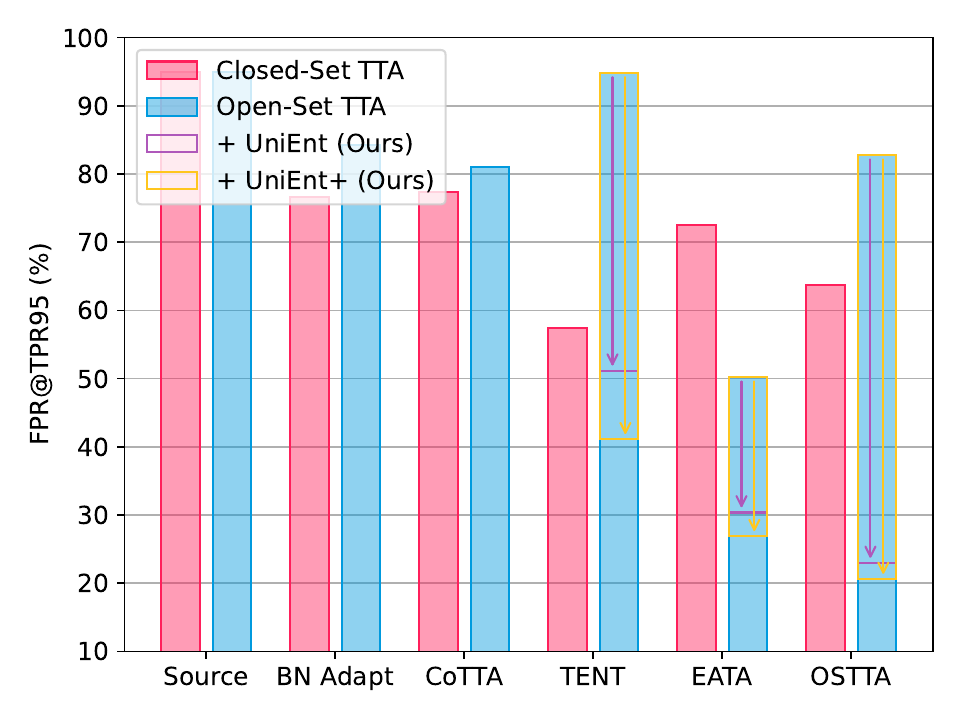}}
    \caption{Performance comparison of existing TTA methods under closed-set and open-set settings.}
    \label{fig:comparison}
\end{figure}

There is no label of the test data to provide supervised imformation during TTA, an entropy minimization or a self-training strategy is widely adopted in existing methods. While previous studies~\cite{zagoruyko2016wide,nado2020evaluating,wang2022continual,wang2021tent,niu2022efficient,lee2023towards} focused on improving the performance of closed-set TTA, we empirically find that they exhibit performance degradation with open-set samples included. As shown in \fref{fig:comparison}, We first compare the performance of existing TTA methods under different settings. Specifically, we conduct closed-set experiments on CIFAR-100-C~\cite{hendrycks2019benchmarking}, \ie, updating the model and measuring the performance of the adapted model with only the test samples from known classes, and the open-set counterparts are extracted from \tref{tab:cifar}. Experimental results show that applying existing methods to open-set TTA leads to the degradation of both the classification performance on known classes and the detection performance on unknown classes. We argue the degradation is caused by the following two reasons. First, the introduce of samples from unknown classes leads to the incorrect estimation of normalization statistics by the model, which results in unreliable updating of the model parameters. Second, entropy minimization-based methods achieved competitive closed-set results by making the model confident on the predictions. However, minimizing entropy on samples from unknown classes destroys the model confidence, which is an undesirable result. We believe that a good model confidence is very important, especially in open-set TTA, because it can tell us how much can we trust the adapted model's predictions.

\subsection{Distribution-aware Filter}

We first model the open-set data distribution as shown in \eref{eq:open}:
\begin{equation}
    \mathcal{P}_\text{OPEN}:=\pi\mathcal{P}_\text{csID}+(1-\pi)\mathcal{P}_\text{csOOD},
    \label{eq:open}
\end{equation}
where $\pi\in[0,1]$. \Eref{eq:open} contains two distributions that the model may encounter during TTA:
\begin{itemize}
    \item Covariate-shifted ID $\mathcal{P}_\text{csID}$ shares the label space with the training data, whereas the input space suffers from style and domain shifts.
    \item Covariate-shifted OOD $\mathcal{P}_\text{csOOD}$ differs from those of the training data in both the label space and the input space.
\end{itemize}

We define the csOOD score for each test sample as:
\begin{equation}
    S(\mathbf{x})=\nu\left(\max_c\frac{g_{\theta_0}(\mathbf{x})\cdot p_c}{\|g_{\theta_0}(\mathbf{x})\|\|p_c\|}\right),
    \label{eq:score}
\end{equation}
where $\nu(\cdot)$ denotes min-max normalization with the range of $[0,1]$, $g_{\theta_0}$ denotes the feature extractor of source domain pre-trained model, $p_c$ denotes the source domain prototype of class $c$.

\begin{figure}[t]
    \centering
    \includegraphics[width=\linewidth]{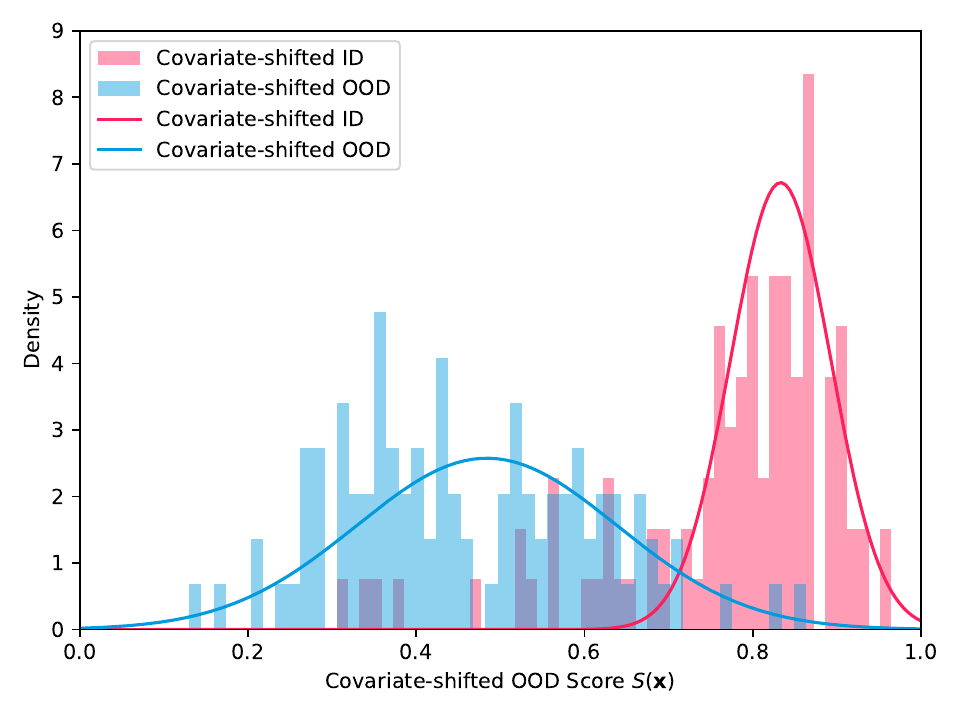}
    \caption{The csOOD score $S(\mathbf{x})$ presents a bimodal distribution.}
    \label{fig:score}
\end{figure}

As shown in \fref{fig:score}, we empirically found that $S(\mathbf{x})$ can distinguish between csID samples and csOOD samples. To be more specific, the distribution of $S(\mathbf{x})$ appears to be bimodal, and its two peaks indicate csID and csOOD modes, respectively. In order to select the optimal threshold, we model the distribution of $S(\mathbf{x})$ as a Gaussian mixture model (GMM) with two components, where the component with larger mean corresponds to the csID samples, and vice versa:
\begin{equation}
    \begin{split}
        \mathcal{P}(\mathbf{x})=&\pi(\mathbf{x})\mathcal{N}(\mathbf{x}\mid\mu_\text{csID},\sigma_\text{csID}^2)\\
        &+(1-\pi(\mathbf{x}))\mathcal{N}(\mathbf{x}\mid\mu_\text{csOOD},\sigma_\text{csOOD}^2)
    \end{split},
\end{equation}
where $\pi(\mathbf{x})$ denotes the probability that $S(\mathbf{x})$ belongs to the csID component, $\mu_\text{csID}$, $\sigma_\text{csID}^2$ and $\mu_\text{csOOD}$, $\sigma_\text{csOOD}^2$ represent the mean and variance of the csID and csOOD components, respectively. Further, $\pi(\mathbf{x})$ can be easily obtained using the EM algorithm.

Then, we can split $\mathcal{B}_t$ into $\mathcal{B}_{t,\text{csID}}$ and $\mathcal{B}_{t,\text{csOOD}}$ through \eref{eq:split}:
\begin{align}
    \begin{split}
        \mathcal{B}_{t,\text{csID}}&=\{\mathbf{x}\mid\mathbf{x}\in\mathcal{B}_t\wedge\pi(\mathbf{x})\geq0.5\}\\
        \mathcal{B}_{t,\text{csOOD}}&=\{\mathbf{x}\mid\mathbf{x}\in\mathcal{B}_t\wedge\pi(\mathbf{x})<0.5\}
    \end{split},
    \label{eq:split}
\end{align}
where $\mathcal{B}_{t,\text{csID}}$ and $\mathcal{B}_{t,\text{csOOD}}$ are the mini-batches of pseudo csID and pseudo csOOD samples at timestamp $t$, respectively.

\subsection{Unified Entropy Optimization}

\paragraph{UniEnt.} Based on the previous sections, we consider minimizing the entropy of the model's predictions of the samples from known classes, which can solve the inaccurate estimation of the data distribution and yield more reliable adaptation. However, the samples from unknown classes have not been explored effectively. Inspired by previous work~\cite{hendrycks2019deep,katz2022training,yang2023auto}, we propose to make the model produce approximately uniform predictions via entropy maximization instead, which can solve the inaccurate estimation of the model confidence and help distinguish known classes samples from unknown classes samples. The overall test-time optimization objective can be written as:
\begin{equation}
    \mathcal{L}_{t,\text{csID}}=\frac{1}{\|\mathcal{B}_{t,\text{csID}}\|}\sum_{\mathbf{x}\in\mathcal{B}_{t,\text{csID}}}H(f_{\theta_t}(\mathbf{x})),
\end{equation}
\begin{equation}
    \mathcal{L}_{t,\text{csOOD}}=\frac{1}{\|\mathcal{B}_{t,\text{csOOD}}\|}\sum_{\mathbf{x}\in\mathcal{B}_{t,\text{csOOD}}}H(f_{\theta_t}(\mathbf{x})),
\end{equation}
\begin{equation}
    \min_{\theta_t}\mathcal{L}_t=\mathcal{L}_{t,\text{csID}}-\lambda_1\mathcal{L}_{t,\text{csOOD}}-\lambda_2H(\bar{f}_{\theta_t}),
    \label{eq:UniEnt}
\end{equation}
where $\lambda_1$ and $\lambda_2$ are trade-off hyperparameters.

\paragraph{UniEnt+.} In the distribution-aware filter, we distinguish csID samples from csOOD samples roughly, which inevitably introduces some noise. To address this problem, we propose a weighting scheme to achieve entropy minimization for known classes and entropy maximization for unknown classes at the same time. The objective can be reformulated as follows:
\begin{equation}
    \begin{split}
        \min_{\theta_t}\mathcal{L}_t=&\frac{1}{\|\mathcal{B}_t\|}\sum_{\mathbf{x}\in\mathcal{B}_t}\pi(\mathbf{x})H(f_{\theta_t}(\mathbf{x}))\\
        &-\lambda_1\frac{1}{\|\mathcal{B}_t\|}\sum_{\mathbf{x}\in\mathcal{B}_t}(1-\pi(\mathbf{x}))H(f_{\theta_t}(\mathbf{x}))\\
        &-\lambda_2H(\bar{f}_{\theta_t})
    \end{split}.
    \label{eq:UniEnt+}
\end{equation}

\section{Experiments}
\label{sec:exp}

\begin{table*}[t]
    \centering
    \resizebox{\linewidth}{!}{%
        \begin{tabular}{l|cccc|cccc|cccc}
            \toprule
            \multirow{2}{*}{Method} & \multicolumn{4}{c|}{CIFAR-10-C} & \multicolumn{4}{c|}{CIFAR-100-C} & \multicolumn{4}{c}{Average} \\ \cmidrule{2-13}
             & Acc$\uparrow$ & AUROC$\uparrow$ & FPR@TPR95$\downarrow$ & OSCR$\uparrow$ & Acc$\uparrow$ & AUROC$\uparrow$ & FPR@TPR95$\downarrow$ & OSCR$\uparrow$ & Acc$\uparrow$ & AUROC$\uparrow$ & FPR@TPR95$\downarrow$ & OSCR$\uparrow$ \\ \midrule
            Source~\cite{zagoruyko2016wide} & 81.73 & 77.89 & 79.45 & 68.44 & 53.25 & 60.55 & 94.98 & 39.87 & 67.49 & 69.22 & 87.22 & 54.16 \\
            BN Adapt~\cite{nado2020evaluating} & 84.20 & 80.40 & 76.84 & 72.13 & 57.16 & 72.45 & 84.29 & 47.10 & 70.68 & 76.43 & 80.57 & 59.62 \\
            CoTTA~\cite{wang2022continual} & 85.77 & 85.89 & 72.40 & 77.26 & 56.46 & 77.04 & 80.96 & 48.95 & 71.12 & 81.47 & 76.68 & 63.11 \\ \midrule
            TENT~\cite{wang2021tent} & 79.38 & 65.39 & 95.94 & 56.73 & 54.74 & 65.00 & 94.79 & 42.24 & 67.06 & 65.20 & 95.37 & 49.49 \\
            \rowcolor[HTML]{EFEFEF} 
            + UniEnt & \best{84.31} (+4.93) & \second{92.28} (+26.89) & \second{36.74} (-59.20) & \second{80.32} (+23.59) & \best{59.07} (+4.33) & \second{89.28} (+24.28) & \second{51.14} (-43.65) & \second{56.26} (+14.02) & \best{71.69} (+4.63) & \second{90.78} (+25.59) & \second{43.94} (-51.43) & \second{68.29} (+18.81) \\
            \rowcolor[HTML]{EFEFEF} 
            + UniEnt+ & \second{84.03} (+4.65) & \best{93.18} (+27.79) & \best{32.74} (-63.20) & \best{80.62} (+23.89) & \second{58.58} (+3.84) & \best{91.39} (+26.39) & \best{41.09} (-53.70) & \best{56.36} (+14.12) & \second{71.31} (+4.25) & \best{92.29} (+27.09) & \best{36.92} (-58.45) & \best{68.49} (+19.01) \\ \midrule
            EATA~\cite{niu2022efficient} & 80.92 & 84.32 & 71.66 & 72.63 & \best{60.63} & 88.64 & 50.18 & 57.24 & 70.78 & 86.48 & 60.92 & 64.94 \\
            \rowcolor[HTML]{EFEFEF} 
            + UniEnt & \second{84.31} (+3.39) & \best{97.15} (+12.83) & \best{13.25} (-58.41) & \second{82.99} (+10.36) & \second{59.75} (-0.88) & \second{93.42} (+4.78) & \second{30.36} (-19.82) & \second{57.99} (+0.75) & \second{72.03} (+1.26) & \second{95.29} (+8.81) & \second{21.81} (-39.12) & \second{70.49} (+5.55) \\
            \rowcolor[HTML]{EFEFEF} 
            + UniEnt+ & \best{85.18} (+4.26) & \second{96.97} (+12.65) & \second{14.28} (-57.38) & \best{83.67} (+11.04) & 59.71(-0.92) & \best{94.23} (+5.59) & \best{26.87} (-23.31) & \best{58.19} (+0.95) & \best{72.45} (+1.67) & \best{95.60} (+9.12) & \best{20.58} (-40.35) & \best{70.93} (+6.00) \\ \midrule
            OSTTA~\cite{lee2023towards} & \best{84.44} & 72.74 & 77.02 & 65.17 & \best{60.03} & 75.37 & 82.75 & 51.35 & \best{72.24} & 74.06 & 79.89 & 58.26 \\
            \rowcolor[HTML]{EFEFEF} 
            + UniEnt & 82.46 (-1.98) & \second{96.20} (+23.46) & \second{16.37} (-60.65) & \second{80.51} (+15.34) & 58.69 (-1.34) & \second{94.84} (+19.47) & \second{22.95} (-59.80) & \second{57.28} (+5.93) & 70.58 (-1.66) & \second{95.52} (+21.47) & \second{19.66} (-60.23) & \second{68.90} (+10.64) \\
            \rowcolor[HTML]{EFEFEF} 
            + UniEnt+ & \second{84.30} (-0.14) & \best{97.38} (+24.64) & \best{11.56} (-65.46) & \best{82.91} (+17.74) & \second{58.93} (-1.10) & \best{95.42} (+20.05) & \best{20.59} (-62.16) & \best{57.69} (+6.34) & \second{71.62} (-0.62) & \best{96.40} (+22.35) & \best{16.08} (-63.81) & \best{70.30} (+12.04) \\ \bottomrule
        \end{tabular}%
    }
    \caption{Results of different methods on CIFAR benchmarks. $\uparrow$ indicates that larger values are better, and vice versa. All values are percentages. The \textbf{bold} values indicate the best results, and the \underline{underlined} values indicate the second best results.}
    \label{tab:cifar}
\end{table*}

\begin{table}[t]
    \centering
    \resizebox{\linewidth}{!}{%
        \begin{tabular}{l|cccc}
            \toprule
            \multirow{2}{*}{Method} & \multicolumn{4}{c}{Tiny-ImageNet-C} \\ \cmidrule{2-5}
             & Acc$\uparrow$ & AUROC$\uparrow$ & FPR@TPR95$\downarrow$ & OSCR$\uparrow$ \\ \midrule
            Source~\cite{zagoruyko2016wide} & 22.29 & 53.79 & 93.41 & 16.29 \\
            BN Adapt~\cite{nado2020evaluating} & 37.00 & 61.06 & 90.90 & 28.50 \\ \midrule
            TENT~\cite{wang2021tent} & 28.96 & 49.78 & 95.96 & 19.02 \\
            \rowcolor[HTML]{EFEFEF} 
            + UniEnt & \second{37.23} (+8.27) & \best{63.92} (+14.14) & \second{89.72} (-6.24) & \best{30.18} (+11.16) \\
            \rowcolor[HTML]{EFEFEF} 
            + UniEnt+ & \best{37.31} (+8.35) & \second{63.83} (+14.05) & \best{89.12} (-6.84) & \second{30.12} (+11.10) \\ \midrule
            EATA~\cite{niu2022efficient} & 37.09 & 57.55 & 93.22 & 27.91 \\
            \rowcolor[HTML]{EFEFEF} 
            + UniEnt & \second{37.54} (+0.45) & \best{64.34} (+6.79) & \best{89.23} (-3.99) & \second{30.59} (+2.68) \\
            \rowcolor[HTML]{EFEFEF} 
            + UniEnt+ & \best{38.65} (+1.56) & \second{62.30} (+4.75) & \second{90.88} (-2.34) & \best{30.95} (+3.04) \\ \midrule
            OSTTA~\cite{lee2023towards} & \best{37.29} & 55.66 & 94.34 & \best{27.74} \\
            \rowcolor[HTML]{EFEFEF} 
            + UniEnt & 33.72 (-3.57) & \best{62.69} (+7.03) & \second{89.67} (-4.67) & 26.63 (-1.11) \\
            \rowcolor[HTML]{EFEFEF} 
            + UniEnt+ & \second{34.47} (-2.82) & \second{61.28} (+5.62) & \best{89.56} (-4.78) & \second{26.65} (-1.09) \\ \bottomrule
        \end{tabular}%
    }
    \caption{Results of different methods on Tiny-ImageNet-C.}
    \label{tab:tiny_imagenet_c}
\end{table}

\begin{table*}[t]
    \centering
    \resizebox{\linewidth}{!}{%
        \begin{tabular}{l|cc|cccc|cccc}
            \toprule
            \multirow{2}{*}{Method} & \multirow{2}{*}{$\mathcal{L}_{t,\text{csID}}$} & \multirow{2}{*}{$\mathcal{L}_{t,\text{csOOD}}$} & \multicolumn{4}{c|}{CIFAR-10-C} & \multicolumn{4}{c}{CIFAR-100-C} \\ \cmidrule{4-11}
             &  &  & Acc$\uparrow$ & AUROC$\uparrow$ & FPR@TPR95$\downarrow$ & OSCR$\uparrow$ & Acc$\uparrow$ & AUROC$\uparrow$ & FPR@TPR95$\downarrow$ & OSCR$\uparrow$ \\ \midrule
            \multirow{3}{*}{TENT~\cite{wang2021tent}} & \textcolor{red}{\ding{55}} & \textcolor{red}{\ding{55}} & 79.38 & 65.39 & 95.94 & 56.73 & 54.74 & 65.00 & 94.79 & 42.24 \\
             & \textcolor{green}{\ding{51}} & \textcolor{red}{\ding{55}} & \best{85.04} (+5.66) & \second{81.80} (+16.41) & \second{68.89} (-27.05) & \second{73.57} (+16.84) & \best{59.30} (+4.56) & \second{86.09} (+21.09) & \second{63.65} (-31.14) & \second{55.55} (+13.31) \\
             & \textcolor{green}{\ding{51}} & \textcolor{green}{\ding{51}} & \second{84.31} (+4.93) & \best{92.28} (+26.89) & \best{36.74} (-59.20) & \best{80.32} (+23.59) & \second{59.07} (+4.33) & \best{89.28} (+24.28) & \best{51.14} (-43.65) & \best{56.26} (+14.02) \\ \midrule
            \multirow{3}{*}{EATA~\cite{niu2022efficient}} & \textcolor{red}{\ding{55}} & \textcolor{red}{\ding{55}} & 80.92 & \second{84.32} & 71.66 & 72.63 & \best{60.63} & \second{88.64} & \second{50.18} & 57.24 \\
             & \textcolor{green}{\ding{51}} & \textcolor{red}{\ding{55}} & \best{85.53} (+4.61) & 82.94 (-1.38) & \second{67.95} (-3.71) & \second{74.85} (+2.22) & \second{60.46} (-0.17) & 88.53 (-0.11) & 54.30 (+4.12) & \second{57.26} (+0.02) \\
             & \textcolor{green}{\ding{51}} & \textcolor{green}{\ding{51}} & \second{84.31} (+3.39) & \best{97.15} (+12.83) & \best{13.25} (-58.41) & \best{82.99} (+10.36) & 59.75 (-0.88) & \best{93.42} (+4.78) & \best{30.36} (-19.82) & \best{57.99} (+0.75) \\ \midrule
            \multirow{3}{*}{OSTTA~\cite{lee2023towards}} & \textcolor{red}{\ding{55}} & \textcolor{red}{\ding{55}} & \second{84.44} & 72.74 & 77.02 & 65.17 & \best{60.03} & 75.37 & 82.75 & 51.35 \\
             & \textcolor{green}{\ding{51}} & \textcolor{red}{\ding{55}} & \best{84.86} (+0.42) & \second{84.96} (+12.22) & \second{62.66} (-14.36) & \second{75.84} (+10.67) & \second{58.95} (-1.08) & \second{90.62} (+15.25) & \second{44.79} (-37.96) & \second{56.50} (+5.15) \\
             & \textcolor{green}{\ding{51}} & \textcolor{green}{\ding{51}} & 82.46 (-1.98) & \best{96.20} (+23.46) & \best{16.37} (-60.65) & \best{80.51} (+15.34) & 58.69 (-1.34) & \best{94.84} (+19.47) & \best{22.95} (-59.80) & \best{57.28} (+5.93) \\ \bottomrule
        \end{tabular}%
    }
    \caption{Ablation study on CIFAR benchmarks. We investigate the effectiveness of $\mathcal{L}_{t,\text{csID}}$ and $\mathcal{L}_{t,\text{csOOD}}$ in \eref{eq:UniEnt} for UniEnt.}
    \label{tab:ablation}
\end{table*}

\begin{table*}[t]
    \centering
    \resizebox{\linewidth}{!}{%
        \begin{tabular}{llccccc}
            \toprule
            \multicolumn{2}{l}{Method} & 0.1 & 0.2 & 0.5 & 1.0 & $\Delta$ \\ \midrule
            \multirow{2}{*}{TENT~\cite{wang2021tent}} & + UniEnt & (59.09, 89.11, 51.68, 56.20) & (59.07, 89.28, 51.14, 56.26) & (58.92, 89.59, 50.16, 56.22) & (58.76, 89.95, 48.92, 56.21) & (0.33, 0.84, 2.76, 0.06) \\
             & + UniEnt+ & (58.64, 91.18, 41.79, 56.34) & (58.58, 91.39, 41.09, 56.36) & (58.41, 91.68, 40.22, 56.33) & (58.12, 91.89, 39.68, 56.13) & (0.52, 0.71, 2.11, 0.23) \\ \midrule
            \multirow{2}{*}{EATA~\cite{niu2022efficient}} & + UniEnt & (59.50, 93.34, 30.72, 57.72) & (59.75, 93.42, 30.36, 57.99) & (59.37, 92.56, 34.98, 57.40) & (59.58, 93.82, 28.29, 57.97) & (0.38, 1.26, 6.69, 0.59) \\
             & + UniEnt+ & (59.73, 93.47, 30.25, 58.00) & (59.81, 93.88, 27.84, 58.17) & (59.71, 94.23, 26.87, 58.19) & (59.62, 93.47, 30.37, 57.91) & (0.19, 0.76, 3.50, 0.28) \\ \midrule
            \multirow{2}{*}{OSTTA~\cite{lee2023towards}} & + UniEnt & (58.85, 93.89, 26.59, 57.14) & (58.82, 94.32, 24.94, 57.24) & (58.69, 94.84, 22.95, 57.28) & (57.88, 94.80, 23.51, 56.51) & (0.97, 0.95, 3.64, 0.77) \\
             & + UniEnt+ & (59.25, 94.19, 24.62, 57.54) & (59.15, 94.84, 22.29, 57.69) & (58.93, 95.42, 20.59, 57.69) & (58.20, 95.65, 20.12, 57.06) & (1.05, 1.46, 4.50, 0.63) \\ \bottomrule
        \end{tabular}%
    }
    \caption{Performance of UniEnt and UniEnt+ with varying $\lambda_1$ on CIFAR-100-C. The values in the table are presented as (Acc, AUROC, FPR@TPR95, OSCR). $\Delta$ is the difference between the maximum and minimum values when $\lambda_1$ take different values. Smaller $\Delta$ values represent better robustness.}
    \label{tab:sensitivity1}
\end{table*}

\begin{table*}[t]
    \centering
    \resizebox{\linewidth}{!}{%
        \begin{tabular}{llccccc}
            \toprule
            \multicolumn{2}{l}{Method} & 0.1 & 0.2 & 0.5 & 1.0 & $\Delta$ \\ \midrule
            \multirow{2}{*}{TENT~\cite{wang2021tent}} & + UniEnt & (59.44, 87.02, 60.32, 55.93) & (59.07, 89.28, 51.14, 56.26) & (58.09, 92.87, 33.24, 56.23) & (56.62, 94.53, 25.26, 55.24) & (2.82, 7.51, 35.06, 1.02) \\
             & + UniEnt+ & (59.19, 87.95, 57.31, 56.04) & (58.58, 91.39, 41.09, 56.36) & (56.71, 94.57, 25.02, 55.34) & (53.13, 94.93, 24.19, 52.01) & (6.06, 6.98, 33.12, 4.35) \\ \midrule
            \multirow{2}{*}{EATA~\cite{niu2022efficient}} & + UniEnt & (60.54, 88.14, 55.48, 57.15) & (60.06, 89.45, 50.99, 57.16) & (59.75, 93.42, 30.36, 57.99) & (58.26, 95.07, 22.18, 57.02) & (2.28, 6.93, 33.30, 0.97) \\
             & + UniEnt+ & (60.35, 89.49, 50.20, 57.44) & (60.51, 91.03, 42.50, 58.02) & (59.71, 94.23, 26.87, 58.19) & (59.03, 95.28, 21.20, 57.81) & (1.48, 5.79, 29.00, 0.75) \\ \midrule
            \multirow{2}{*}{OSTTA~\cite{lee2023towards}} & + UniEnt & (58.69, 94.84, 22.95, 57.28) & (56.63, 95.43, 21.02, 55.46) & (49.85, 93.77, 32.12, 48.59) & (43.89, 91.19, 47.50, 42.41) & (14.80, 4.24, 26.48, 14.87) \\
             & + UniEnt+ & (59.15, 94.84, 22.29, 57.69) & (57.55, 95.82, 18.91, 56.43) & (50.31, 94.09, 30.05, 49.11) & (43.66, 91.78, 43.35, 42.28) & (15.49, 4.04, 24.44, 15.41) \\ \bottomrule
        \end{tabular}%
    }
    \caption{Performance of UniEnt and UniEnt+ with varying $\lambda_2$ on CIFAR-100-C. $\Delta$ is the difference between the maximum and minimum values when $\lambda_2$ take different values.}
    \label{tab:sensitivity2}
\end{table*}

\subsection{Setup}

\paragraph{Datasets.} Following previous studies, we evaluate our proposed methods on the widely used corruption benchmark datasets: CIFAR-10-C, CIFAR-100-C, and Tiny-ImageNet-C~\cite{hendrycks2019benchmarking}. Each dataset contains 15 types of corruptions with 5 severity levels, all our experiments are conducted under the most severe corruption level 5. Pre-trained models are trained on the clean training set and tested and adapted on the corrupted test set. Following OSTTA~\cite{lee2023towards}, we apply the same corruption type to the original SVHN~\cite{netzer2011reading} and ImageNet-O~\cite{hendrycks2021natural} test sets to generate the SVHN-C and ImageNet-O-C datasets. We use SVHN-C and ImageNet-O-C as the covariate shifted OOD datasets for CIFAR-10/100-C and Tiny-ImageNet-C, respectively.

\paragraph{Evaluation protocols.} Following recent research~\cite{wang2022continual,wang2021tent,niu2022efficient,lee2023towards}, we evaluate TTA methods under continuously changing domains without resetting the parameters after each domain. At test time, the corrupted images are provided to the model in an online fashion. After encountering a mini-batch of test data, the model makes predictions and updates parameters immediately. The predictions of test data arriving at timestamp $t$ will not be affected by any test data arriving after timestamp $t$. We construct the mini-batch using the same number of csID samples and csOOD samples. Regarding the model's adaptation performance on csID data, we use the accuracy metric. To evaluate whether the adapted model can detect csOOD data robustly, we measure the area under the receiver operating characteristic curve (AUROC) and the false positive rate of csOOD samples when the true positive rate of csID samples is at 95\% (FPR@TPR95). As we pursue a good trade-off between the classification accuracy on csID data and the detection accuracy on csOOD data, we also report the open-set classification rate (OSCR)~\cite{dhamija2018reducing} to measure the balanced performance.

\paragraph{Baseline methods.} We mainly compare our method with two types of pervious methods in TTA: 1) entropy-free methods: \textbf{Source} directly evaluates the test data using the source model without adaptation. \textbf{BN Adapt}~\cite{nado2020evaluating} updates batch normalization statistics with the test data during TTA. \textbf{CoTTA}~\cite{wang2022continual} adopts the teacher-student architecture to provide weight-averaged and augmentation-averaged pseudo-labels to reduce error accumulation, combined with stochastic restoration to avoid catastrophic forgetting. 2) entropy-based methods: \textbf{TENT}~\cite{wang2021tent} estimates normalization statistics and optimizes channel-wise affine transformations through entropy minimization. \textbf{EATA}~\cite{niu2022efficient} selects reliable and non-redundant samples for model adaptation, the former achieve prediction entropy lower than a pre-defined threshold and the latter have diverse model outputs. In addition, the fisher regularization is introduced to prevent catastrophic forgetting. \textbf{OSTTA}~\cite{lee2023towards} uses the wisdom of crowds to filter out the samples with lower confidence values in the adapted model than in the original model. Our methods can be easily applied to existing entropy-based methods without additional modification. Regarding applying our methods to EATA and OSTTA, we apply the filtering methods and keep everything else the same.

\paragraph{Implementation details.} For experiments on CIFAR benchmarks, following pervious studies~\cite{choi2022improving,lim2023ttn,lee2023towards}, we use the WideResNet~\cite{zagoruyko2016wide} with 40 layers and widen factor of 2. The model pre-trained with AugMix~\cite{hendrycks2020augmix} is available from RobustBench~\cite{croce2021robustbench}. For Tiny-ImageNet-C, we pre-train ResNet50~\cite{he2016deep} on the Tiny-ImageNet~\cite{le2015tiny} training set, as OSTTA~\cite{lee2023towards} did. The model is initialized with the pre-trained weights on ImageNet~\cite{deng2009imagenet} and optimized for 50 epochs using SGD~\cite{ruder2016overview} with a batch size of 256. The initial learning rate is set to 0.01 and adjust using a cosine annealing schedule. During TTA, we use Adam~\cite{kingma2014adam} optimizer with the batch size of 200 for all experiments. The learning rate is set to 0.001 and 0.01 for entropy-based methods (TENT~\cite{wang2021tent}, EATA~\cite{niu2022efficient}, OSTTA) and CoTTA~\cite{wang2022continual}, respectively. We use the energy score~\cite{liu2020energy} to measure the ability of the adapted model to detect unknown classes. Furthermore, following T3A~\cite{iwasawa2021test}, we use the weights of the linear classifier as the source domain prototypes, and thus our approach is source-free. Entropy-based methods update only the affine parameters, while CoTTA updates all parameters.

\subsection{Results}

\paragraph{CIFAR benchmarks.} We first conduct experiments on the most common CIFAR benchmarks, and the results are presented in \tref{tab:cifar}. From \tref{tab:cifar}, we can see that UniEnt and UniEnt+ significantly improve the performance of three different existing TTA methods. For example, on CIFAR-10-C, UniEnt improves the Acc, AUROC, FPR@TPR95 and OSCR of TENT~\cite{wang2021tent} by 4.93\%, 26.89\%, 59.20\% and 23.59\% respectively, while UniEnt+ improves the Acc, AUROC, FPR@TPR95 and OSCR of TENT by 4.65\%, 27.79\%, 63.20\% and 23.89\% respectively.

In more detail, we can observe that TENT~\cite{wang2021tent} and OSTTA~\cite{lee2023towards} perform even worse than Source and BN methods that do not update model parameters in some cases (OSCR decreases by 3.27\%$\sim$15.40\%), which indicates that some existing TTA methods cannot effectively update model parameters with open-set classes included. This can be attributed to the fact that these methods ignore the distribution variations introduced by open-set samples, resulting in the unreliable estimation of normalization statistics and model confidence.

\paragraph{Tiny-ImageNet-C.} We then conduct experiments on a more challenging dataset Tiny-ImageNet-C, and the results are summarized in \tref{tab:tiny_imagenet_c}. As shown in \tref{tab:tiny_imagenet_c}, consistent with previous analysis, UniEnt and UniEnt+ still achieve better performance. Numerically, UniEnt improves the Acc, AUROC, FPR@TPR95 and OSCR of TENT~\cite{wang2021tent} by 8.27\%, 14.14\%, 6.24\% and 11.16\% respectively, while UniEnt+ improves the Acc, AUROC, FPR@TPR95 and OSCR of TENT by 8.35\%, 14.05\%, 6.84\% and 11.10\% respectively.

\subsection{Analysis}

\paragraph{Ablation study.} To verify the effectiveness of different components in $\mathcal{L}_t$ (\eref{eq:UniEnt}), we conduct extensive ablation studies on CIFAR benchmarks. The results are summarized in \tref{tab:ablation}. Compared with the baselines without $\mathcal{L}_{t,\text{csID}}$ and $\mathcal{L}_{t,\text{csOOD}}$ (the same as TENT~\cite{wang2021tent}, EATA~\cite{niu2022efficient} and OSTTA~\cite{lee2023towards}), introducing $\mathcal{L}_{t,\text{csID}}$ improves the classification accuracy of known classes, which indicates that our proposed distribution-aware filter can well distinguish the samples of known classes from the samples of unknown classes. It is worth noting that the introduction of $\mathcal{L}_{t,\text{csID}}$ also leads to better detection performance of unknown classes, which is consistent with the findings obtained in a recent study~\cite{vaze2022open}. With the addition of $\mathcal{L}_{t,\text{csOOD}}$, the model's detection performance of unknown classes has been further improved. Considering the trade-off between the two, UniEnt achieves the optimal OSCR values in most cases.

\paragraph{Hyperparameter sensitivity.} We perform sensitivity analyses on the hyperparameters $\lambda_1$ and $\lambda_2$, as summarized in \tref{tab:sensitivity1} and \tref{tab:sensitivity2}. We first investigate the effect of $\lambda_1$ on CIFAR-100-C, with $\lambda_1$ taking values from $\{0.1,0.2,0.5,1.0\}$ and $\lambda_2$ holds constant. The experimental results show that our methods are robust to the value of $\lambda_1$, the gaps between the best and worst values of Acc, AUROC, FPR@TPR95 and OSCR are 1.05\%, 1.46\%, 6.69\% and 0.77\%, respectively. We then examine how $\lambda_2$ affects csID classification and csOOD detection, with $\lambda_2$ taking values from $\{0.1,0.2,0.5,1.0\}$ and $\lambda_1$ holds constant. It is easy to conclude from the results that a larger $\lambda_2$ leads to better csOOD detection performance, yet at the same time, it may lose some of the csID classification performance, and vice versa. Numerically, different values of $\lambda_2$ will result in the maximum performance differences of 15.49\%, 7.51\%, 35.06\% and 15.41\% for Acc, AUROC, FPR@TPR95 and OSCR, respectively.

\paragraph{Performance under different number of unknown classes.} The number of unknown classes is an important measure representing the complexity of the open-set. We examine the impact of different numbers of unknown classes. Specifically, we perform experiments on the CIFAR-10-C dataset and control the number of unknown classes to vary from 2 to 10, keeping the number of samples constant. From \tref{tab:openness}, we can see that TENT~\cite{wang2021tent} fluctuates with different number of classes while the proposed UniEnt and UniEnt+ are more robust to different number of unknown classes.

\begin{table}[t]
    \centering
    \resizebox{\linewidth}{!}{%
        \begin{tabular}{lcccccc}
            \toprule
            Method & 2 & 4 & 6 & 8 & 10 & $\Delta$ \\ \midrule
            Source~\cite{zagoruyko2016wide} & 70.84 & 69.28 & 69.32 & 69.18 & 68.44 & 2.40 \\
            BN Adapt~\cite{nado2020evaluating} & 72.56 & 72.48 & 72.52 & 72.44 & 72.14 & 0.42 \\ \midrule
            TENT~\cite{wang2021tent} & 49.51 & 48.29 & 51.74 & 49.53 & 50.97 & 3.45 \\
            \rowcolor[HTML]{EFEFEF} 
            + UniEnt & 78.71 & 78.39 & 78.28 & 78.13 & 77.82 & 0.89 \\
            \rowcolor[HTML]{EFEFEF} 
            + UniEnt+ & 78.65 & 78.23 & 78.23 & 78.07 & 77.68 & 0.97 \\ \bottomrule
        \end{tabular}%
    }
    \caption{OSCR of UniEnt and UniEnt+ on CIFAR-10-C under different number of unknown classes.}
    \label{tab:openness}
\end{table}

\paragraph{Performance under different ratios of csOOD to csID samples.} We also perform experiments with different ratios of the number of csOOD samples to the number of csID samples, and the results are displayed in \tref{tab:ratio}. We vary the data ratio from 0.2 to 1.0. It can be observed that our proposed methods are insensitive to the variation of the data ratio while TENT~\cite{wang2021tent} is more sensitive, and thus can be applied to different data ratio cases.

\begin{table}[t]
    \centering
    \resizebox{\linewidth}{!}{%
        \begin{tabular}{lcccccc}
            \toprule
            Method & 0.2 & 0.4 & 0.6 & 0.8 & 1.0 & $\Delta$ \\ \midrule
            Source~\cite{zagoruyko2016wide} & 40.00 & 40.03 & 39.98 & 39.92 & 39.87 & 0.16 \\
            BN Adapt~\cite{nado2020evaluating} & 49.91 & 49.55 & 48.92 & 47.97 & 47.10 & 2.81 \\ \midrule
            TENT~\cite{wang2021tent} & 47.68 & 44.12 & 44.06 & 42.90 & 42.16 & 5.52 \\
            \rowcolor[HTML]{EFEFEF} 
            + UniEnt & 56.84 & 57.48 & 57.13 & 56.77 & 56.26 & 1.22 \\
            \rowcolor[HTML]{EFEFEF} 
            + UniEnt+ & 57.15 & 57.59 & 57.24 & 56.88 & 56.33 & 1.26 \\ \bottomrule
        \end{tabular}%
    }
    \caption{OSCR of UniEnt and UniEnt+ on CIFAR-100-C under different ratios of csOOD to csID samples.}
    \label{tab:ratio}
\end{table}

\begin{figure}[t]
    \centering
    \subfloat[TENT~\cite{wang2021tent}]{\includegraphics[width=0.48\linewidth]{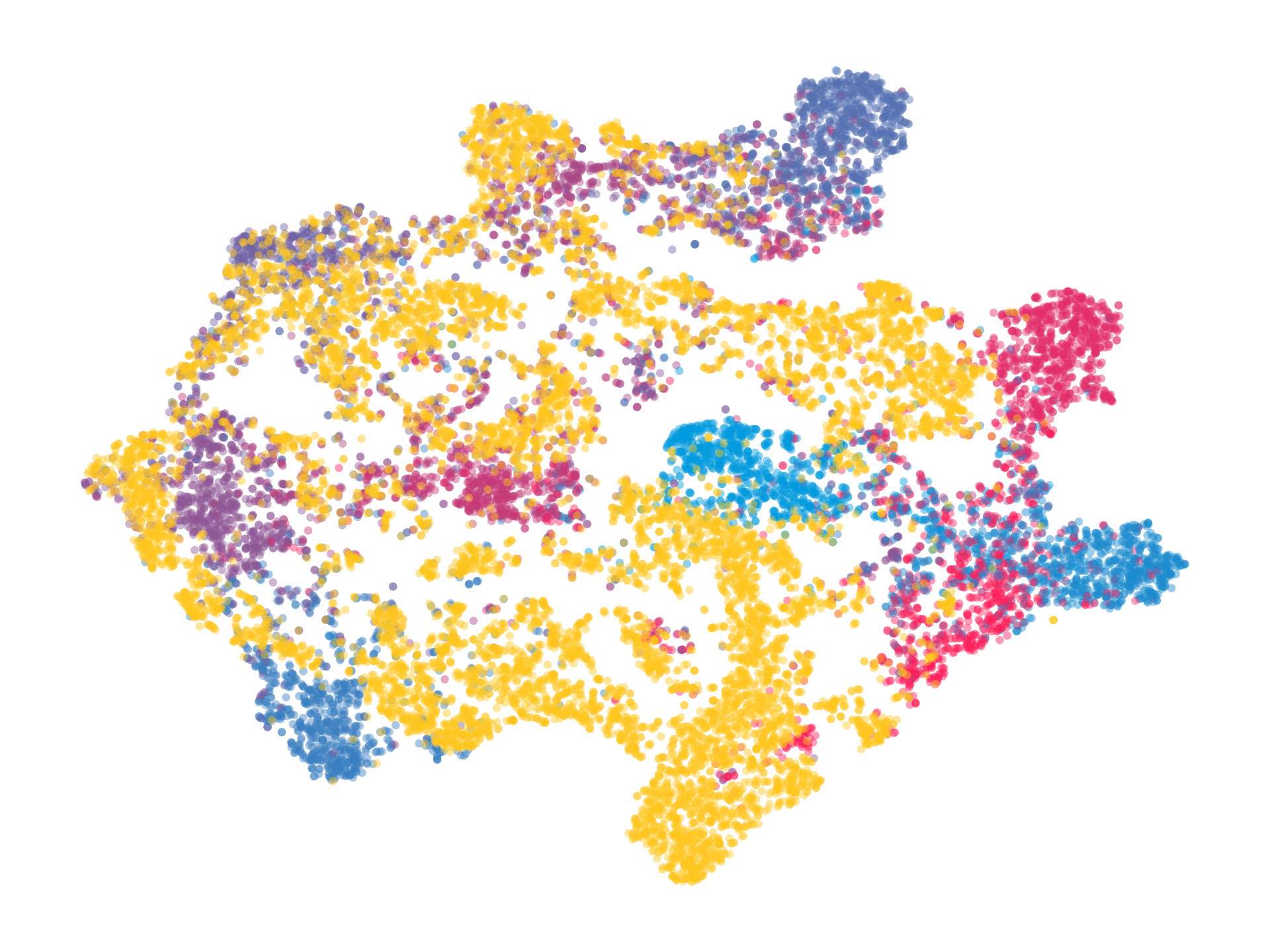}}
    \subfloat[TENT~\cite{wang2021tent} w/ $\mathcal{L}_{t,\text{csID}}$]{\includegraphics[width=0.48\linewidth]{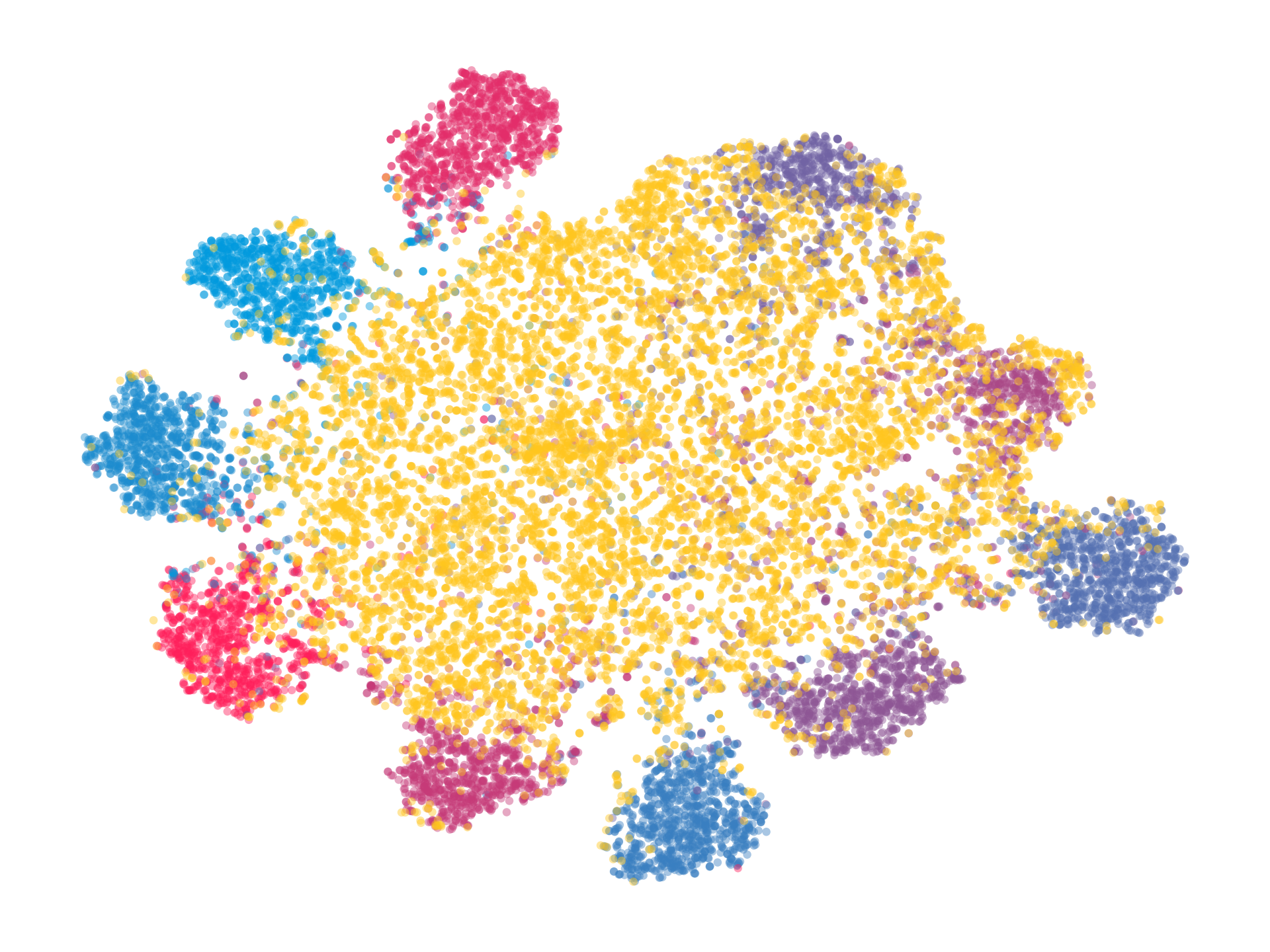}} \\
    \subfloat[TENT~\cite{wang2021tent} w/ UniEnt]{\includegraphics[width=0.48\linewidth]{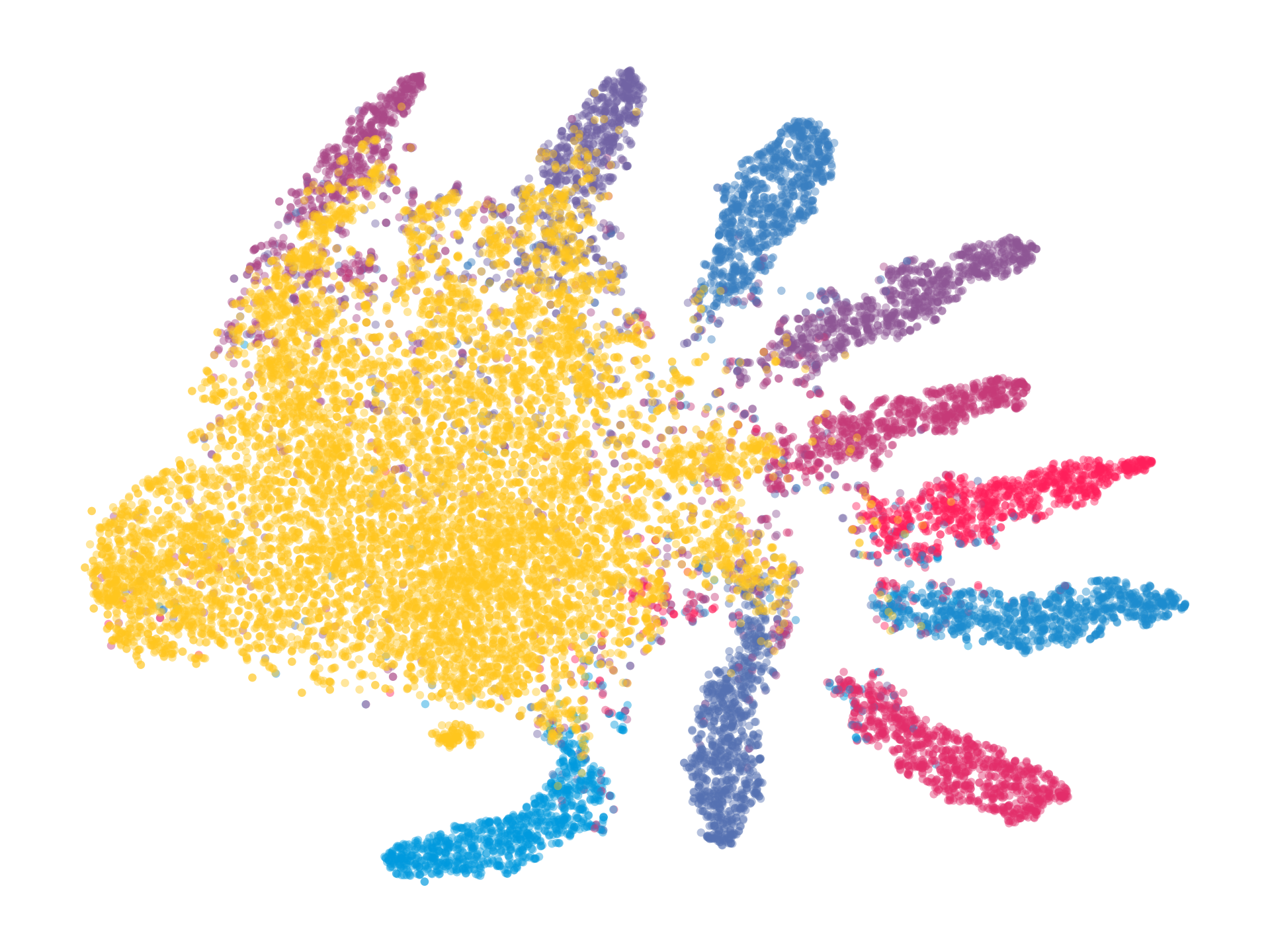}}
    \subfloat[TENT~\cite{wang2021tent} w/ UniEnt+]{\includegraphics[width=0.48\linewidth]{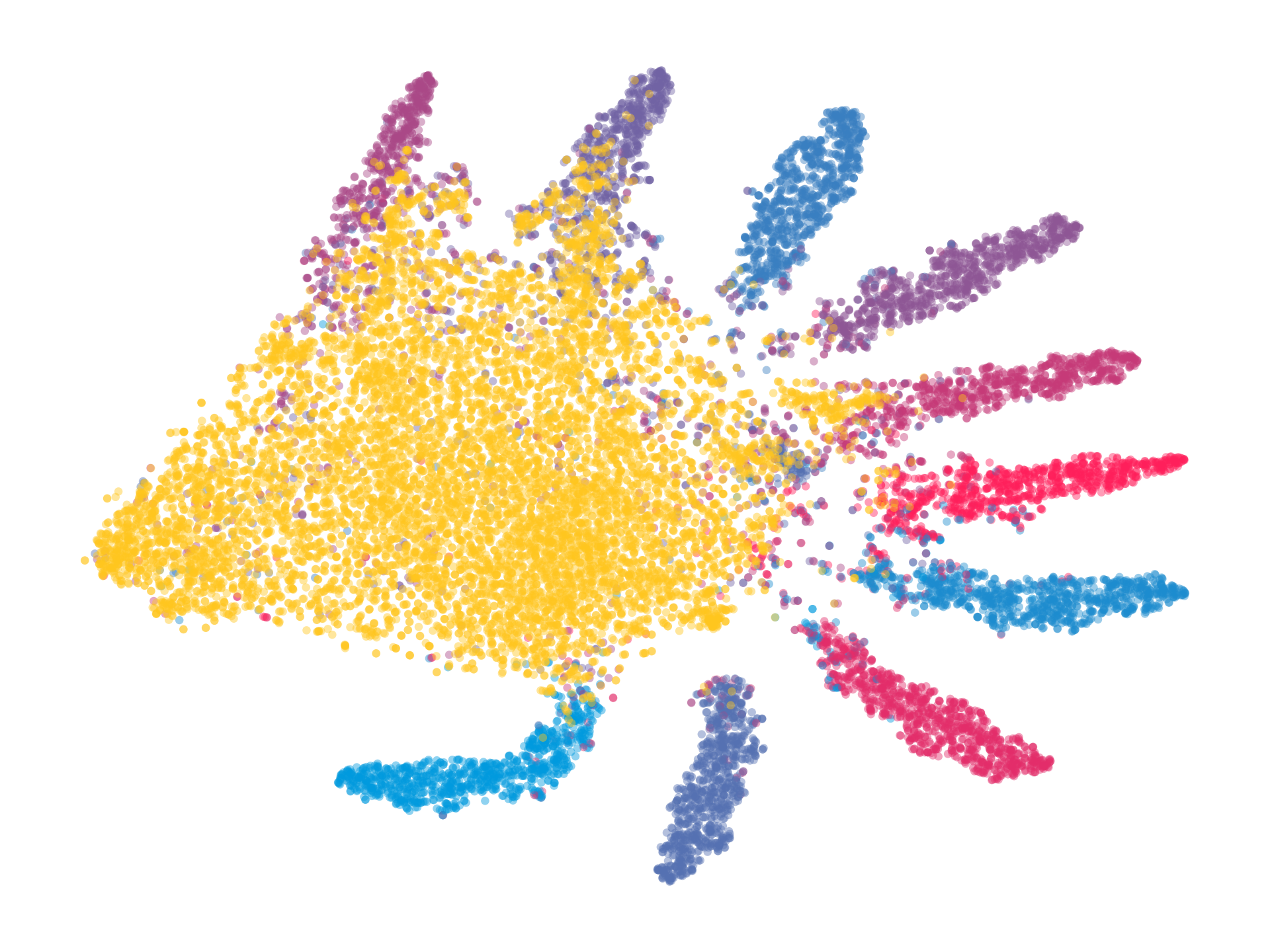}}
    \caption{T-SNE visualization on CIFAR-10-C test set with SVHN-C as csOOD. \textcolor{color1!70}{red} $\to$ \textcolor{color2!70}{blue} denotes csID samples and \textcolor{color3!70}{yellow} denotes csOOD samples.}
    \label{fig:tsne}
\end{figure}

\paragraph{T-SNE visualization.} To illustrate the effects of different methods on csID classification and csOOD detection, we visualize the feature representations of CIFAR-10-C test samples with SVHN-C test samples as csOOD samples via T-SNE~\cite{van2008visualizing} in \fref{fig:tsne}. It can be observed that the features from known classes and unknown classes adapted by TENT~\cite{wang2021tent} are mixed together, while UniEnt and UniEnt+ can better separate them. Furthremore, we observe that filtering out csOOD samples (w/ $\mathcal{L}_{t,\text{csID}}$) can not only improve the classification performance on known classes, but also the detection performance on unknown classes.

\section{Conclusion}

This paper presents a unified entropy optimization framework for open-set test-time adaptation that can be flexibly applied to various existing TTA methods. We first delve into the performance of existing methods under open-set TTA setting, and attribute the performance degradation to the unreliable estimation of normalization statistics and model confidence. To address the above issues, we then propose a distribution-aware filter to preliminary distinguish csID samples from csOOD samples, followed by entropy minimization on csID samples and entropy maximization on csOOD samples. In addition, we propose to leverage sample-level confidence to reduce the noise from hard data partition. Extensive experiments reveal that our methods outperform state-of-the-art TTA methods in open-set scenarios. We hope that more studies can focus on the robustness of TTA methods under open-set, which can facilitate the application of these methods in real scenarios.

\paragraph{Acknowledgements.} This work has been supported by the National Science and Technology Major Project (2022ZD0116500), National Natural Science Foundation of China (U20A20223, 62222609, 62076236), CAS Project for Young Scientists in Basic Research (YSBR-083), and Key Research Program of Frontier Sciences of CAS (ZDBS-LY-7004).

{
    \small
    \bibliographystyle{ieeenat_fullname}
    \bibliography{main}
}

\clearpage
\setcounter{page}{1}
\maketitlesupplementary

\section{Pseudo Code}
\label{sec:pseudo code}

For a better understanding of our proposed methods, we summarize UniEnt and UniEnt+ as Algorithm~\ref{alg:UniEnt} and Algorithm~\ref{alg:UniEnt+}, respectively.

\begin{algorithm}
    \KwIn{Source model $f_{\theta_0}$ pre-trained on the source domain dataset, testing samples $\mathcal{B}_t=\{\mathbf{x}\},t=1,\cdots,T$.}
    \For{$t\leftarrow1$ \KwTo $T$}{
        \For{$\mathbf{x}\in\mathcal{B}_t$}{
            Compute csOOD score for each testing sample via \eref{eq:score}\;
        }
        Obtain $\pi(x)$ via the EM algorithm\;
        Split $\mathcal{B}_t$ into $\mathcal{B}_{t,\text{csID}}$ and $\mathcal{B}_{t,\text{csOOD}}$ via \eref{eq:split}\;
        Update model via \eref{eq:UniEnt}\;
    }
    \KwOut{The predictions $\mathop{\arg\max}_cf_{\theta_t}(\mathbf{x})$ for all $\mathbf{x}\in\mathcal{B}_t,t=1,\cdots,T$.}
    \caption{UniEnt}
    \label{alg:UniEnt}
\end{algorithm}

\begin{algorithm}
    \KwIn{Source model $f_{\theta_0}$ pre-trained on the source domain dataset, testing samples $\mathcal{B}_t=\{\mathbf{x}\},t=1,\cdots,T$.}
    \For{$t\leftarrow1$ \KwTo $T$}{
        \For{$\mathbf{x}\in\mathcal{B}_t$}{
            Compute csOOD score for each testing sample via \eref{eq:score}\;
        }
        Obtain $\pi(x)$ via the EM algorithm\;
        Update model via \eref{eq:UniEnt+}\;
    }
    \KwOut{The predictions $\mathop{\arg\max}_cf_{\theta_t}(\mathbf{x})$ for all $\mathbf{x}\in\mathcal{B}_t,t=1,\cdots,T$.}
    \caption{UniEnt+}
    \label{alg:UniEnt+}
\end{algorithm}

\section{More Analysis}
\label{sec:more analysis}

\paragraph{Scalability of large-scale datasets.}
To demonstrate that our methods can be used for large-scale datasets, we conduct experiments on ImageNet-C~\cite{hendrycks2019benchmarking}. Specifically, we use ResNet-50~\cite{he2016deep} pre-trained with AugMix~\cite{hendrycks2020augmix}, the weights of which can be obtained from RobustBench~\cite{croce2021robustbench}. For optimization, we use the SGD optimizer~\cite{ruder2016overview} with the learning rate of 0.00025 and the batch size of 64. We apply common corruptions and perturbations to ImageNet-O~\cite{hendrycks2021natural} through the official code of~\cite{hendrycks2019benchmarking} to construct ImageNet-O-C as csOOD data. From \Tref{tab:imagenet}, we can see that UniEnt and UniEnt+ consistently improve the performance of the existing baseline methods in the open-set setting.

\begin{table}[t]
    \centering
    \resizebox{\linewidth}{!}{%
        \begin{tabular}{l|cccc}
            \toprule
            \multirow{2}{*}{Method} & \multicolumn{4}{c}{ImageNet-C} \\ \cmidrule{2-5}
             & Acc$\uparrow$ & AUROC$\uparrow$ & FPR@TPR95$\downarrow$ & OSCR$\uparrow$ \\ \midrule
            Source~\cite{zagoruyko2016wide} & 28.21 & 49.63 & 94.74 & 19.81 \\
            BN Adapt~\cite{nado2020evaluating} & 43.57 & 55.89 & 93.39 & 30.42 \\
            CoTTA~\cite{wang2022continual} & 47.67 & 55.58 & 94.51 & 33.80 \\ \midrule
            TENT~\cite{wang2021tent} & 45.82 & 51.34 & 96.47 & 30.33 \\
            \rowcolor[HTML]{EFEFEF} 
            + UniEnt & \best{47.53} (+1.71) & \best{56.33} (+4.99) & \second{95.21} (-1.26) & \best{34.42} (+4.09) \\
            \rowcolor[HTML]{EFEFEF} 
            + UniEnt+ & \second{46.87} (+1.05) & \second{55.86} (+4.52) & \best{95.10} (-1.37) & \second{33.73} (+3.40) \\ \midrule
            EATA~\cite{niu2022efficient} & \second{51.40} & 53.10 & 95.18 & 34.87 \\
            \rowcolor[HTML]{EFEFEF} 
            + UniEnt & 49.60 (-1.80) & \second{58.29} (+5.19) & \second{93.63} (-1.55) & \second{36.28} (+1.41) \\
            \rowcolor[HTML]{EFEFEF} 
            + UniEnt+ & \best{51.57} (+0.17) & \best{59.45} (+6.35) & \best{93.60} (-1.58) & \best{38.27} (+3.40) \\ \midrule
            OSTTA~\cite{lee2023towards} & \second{47.91} & 52.93 & 96.15 & 32.77 \\
            \rowcolor[HTML]{EFEFEF} 
            + UniEnt & \best{47.92} (+0.01) & \best{56.02} (+3.09) & \second{95.23} (-0.92) & \best{34.47} (+1.70) \\
            \rowcolor[HTML]{EFEFEF} 
            + UniEnt+ & 47.47 (-0.44) & \second{55.67} (+2.74) & \best{95.16} (-0.99) & \second{34.03} (+1.26) \\ \bottomrule
        \end{tabular}%
    }
    \caption{Results of different methods on ImageNet-C. $\uparrow$ indicates that larger values are better, and vice versa. All values are percentages. The \best{bold} values indicate the best results, and the \second{underlined} values indicate the second best results. The values in parentheses indicate the improvements of our methods over the baseline methods.}
    \label{tab:imagenet}
\end{table}

\paragraph{Scalability of model architecture.}
Recently, Vision Transformer (ViT)~\cite{dosovitskiy2021image} has demonstrated better performance than Convolutional Neural Network (CNN), we also perform experiments with ViT backbone on ImageNet-C. Specifically, we use DeiT-Base~\cite{touvron2021training} designed in~\cite{tian2022deeper}, which proposes many techniques in the training phase to improve the robustness of the model to common corruptions. The pre-trained weights are also available from RobustBench. We update the affine parameters of the model's layer normalization. \Tref{tab:arch} shows that our approaches are compatible with ViT.

\begin{table*}[t]
    \centering
    \resizebox{\linewidth}{!}{%
        \begin{tabular}{l|cccc|cccc}
            \toprule
            \multirow{2}{*}{Method} & \multicolumn{4}{c|}{ResNet-50} & \multicolumn{4}{c}{DeiT Base} \\ \cmidrule{2-9}
             & Acc$\uparrow$ & AUROC$\uparrow$ & FPR@TPR95$\downarrow$ & OSCR$\uparrow$ & Acc$\uparrow$ & AUROC$\uparrow$ & FPR@TPR95$\downarrow$ & OSCR$\uparrow$ \\ \midrule
            Source~\cite{zagoruyko2016wide} & 28.21 & 49.63 & 94.74 & 19.81 & 56.59 & 56.01 & 91.55 & 36.13 \\
            CoTTA~\cite{wang2022continual} & 47.67 & 55.58 & 94.51 & 33.80 & 60.73 & 53.51 & 93.14 & 37.33 \\ \midrule
            TENT~\cite{wang2021tent} & 45.82 & 51.34 & 96.47 & 30.33 & \best{62.85} & 59.51 & 93.47 & 43.52 \\
            \rowcolor[HTML]{EFEFEF} 
            + UniEnt & \best{47.53} (+1.71) & \best{56.33} (+4.99) & \second{95.21} (-1.26) & \best{34.42} (+4.09) & \second{58.81} (-4.04) & \best{67.10} (+7.59) & \second{90.90} (-2.57) & \best{47.40} (+3.88) \\
            \rowcolor[HTML]{EFEFEF} 
            + UniEnt+ & \second{46.87} (+1.05) & \second{55.86} (+4.52) & \best{95.10} (-1.37) & \second{33.73} (+3.40) & 58.40 (-4.45) & \second{66.69} (+7.18) & \best{90.43} (-3.04) & \second{46.74} (+3.22) \\ \midrule
            EATA~\cite{niu2022efficient} & \second{51.40} & 53.10 & 95.18 & 34.87 & \best{65.38} & 57.95 & 92.92 & 44.29 \\
            \rowcolor[HTML]{EFEFEF} 
            + UniEnt & 49.60 (-1.80) & \second{58.29} (+5.19) & \second{93.63} (-1.55) & \second{36.28} (+1.41) & 59.36 (-6.02) & \best{67.22} (+9.27) & \second{91.63} (-1.29) & \second{48.23} (+3.94) \\
            \rowcolor[HTML]{EFEFEF} 
            + UniEnt+ & \best{51.57} (+0.17) & \best{59.45} (+6.35) & \best{93.60} (-1.58) & \best{38.27} (+3.40) & \second{61.50} (-3.88) & \second{66.96} (+9.01) & \best{89.99} (-2.93) & \best{48.79} (+4.50) \\ \midrule
            OSTTA~\cite{lee2023towards} & \second{47.91} & 52.93 & 96.15 & 32.77 & \best{60.19} & 60.69 & 92.42 & 43.19 \\
            \rowcolor[HTML]{EFEFEF} 
            + UniEnt & \best{47.92} (+0.01) & \best{56.02} (+3.09) & \second{95.23} (-0.92) & \best{34.47} (+1.70) & \second{58.73} (-1.46) & \best{67.62} (+6.93) & \second{90.51} (-1.91) & \best{47.64} (+4.45) \\
            \rowcolor[HTML]{EFEFEF} 
            + UniEnt+ & 47.47 (-0.44) & \second{55.67} (+2.74) & \best{95.16} (-0.99) & \second{34.03} (+1.26) & 58.72 (-1.47) & \second{67.28} (+6.59) & \best{90.02} (-2.40) & \second{47.32} (+4.13) \\ \bottomrule
        \end{tabular}%
    }
    \caption{Results of different methods on ImageNet-C using diverse architectures.}
    \label{tab:arch}
\end{table*}

\paragraph{Performance under long-term open-set test-time adaptation.}
Models deployed in real-world scenarios are exposed to test samples for long periods and need to make reliable predictions at any time. Recent work~\cite{press2023rdumb,lee2023towards} points out that most existing TTA methods perform poorly in long-term settings, even worse than non-updating models. Following~\cite{lee2023towards}, we simulate long-term TTA by repeating adaptation for 10 rounds. During adaptation, the domain changes continuously and the model is never reset. The results are summarized in \Tref{tab:long-term}. We observe that in most cases the performance degradation of our methods is very slight compared to the baseline methods.

\begin{table*}[t]
    \centering
    \subfloat[Results after 1 round of adaptation (\ie, short-term test-time adaptation).]{
        \resizebox{\linewidth}{!}{%
            \begin{tabular}{l|cccc|cccc|cccc}
                \toprule
                \multirow{2}{*}{Method} & \multicolumn{4}{c|}{CIFAR-10-C} & \multicolumn{4}{c|}{CIFAR-100-C} & \multicolumn{4}{c}{Average} \\ \cmidrule{2-13}
                 & Acc$\uparrow$ & AUROC$\uparrow$ & FPR@TPR95$\downarrow$ & OSCR$\uparrow$ & Acc$\uparrow$ & AUROC$\uparrow$ & FPR@TPR95$\downarrow$ & OSCR$\uparrow$ & Acc$\uparrow$ & AUROC$\uparrow$ & FPR@TPR95$\downarrow$ & OSCR$\uparrow$ \\ \midrule
                Source~\cite{zagoruyko2016wide} & 81.73 & 77.89 & 79.45 & 68.44 & 53.25 & 60.55 & 94.98 & 39.87 & 67.49 & 69.22 & 87.22 & 54.16 \\
                BN Adapt~\cite{nado2020evaluating} & 84.20 & 80.40 & 76.84 & 72.13 & 57.16 & 72.45 & 84.29 & 47.10 & 70.68 & 76.43 & 80.57 & 59.62 \\
                CoTTA~\cite{wang2022continual} & 85.77 & 85.89 & 72.40 & 77.26 & 56.46 & 77.04 & 80.96 & 48.95 & 71.12 & 81.47 & 76.68 & 63.11 \\ \midrule
                TENT~\cite{wang2021tent} & 79.38 & 65.39 & 95.94 & 56.73 & 54.74 & 65.00 & 94.79 & 42.24 & 67.06 & 65.20 & 95.37 & 49.49 \\
                \rowcolor[HTML]{EFEFEF} 
                + UniEnt & \best{84.31} (+4.93) & \second{92.28} (+26.89) & \second{36.74} (-59.20) & \second{80.32} (+23.59) & \best{59.07} (+4.33) & \second{89.28} (+24.28) & \second{51.14} (-43.65) & \second{56.26} (+14.02) & \best{71.69} (+4.63) & \second{90.78} (+25.59) & \second{43.94} (-51.43) & \second{68.29} (+18.81) \\
                \rowcolor[HTML]{EFEFEF} 
                + UniEnt+ & \second{84.03} (+4.65) & \best{93.18} (+27.79) & \best{32.74} (-63.20) & \best{80.62} (+23.89) & \second{58.58} (+3.84) & \best{91.39} (+26.39) & \best{41.09} (-53.70) & \best{56.36} (+14.12) & \second{71.31} (+4.25) & \best{92.29} (+27.09) & \best{36.92} (-58.45) & \best{68.49} (+19.01) \\ \midrule
                EATA~\cite{niu2022efficient} & 80.92 & 84.32 & 71.66 & 72.63 & \best{60.63} & 88.64 & 50.18 & 57.24 & 70.78 & 86.48 & 60.92 & 64.94 \\
                \rowcolor[HTML]{EFEFEF} 
                + UniEnt & \second{84.31} (+3.39) & \best{97.15} (+12.83) & \best{13.25} (-58.41) & \second{82.99} (+10.36) & \second{59.75} (-0.88) & \second{93.42} (+4.78) & \second{30.36} (-19.82) & \second{57.99} (+0.75) & \second{72.03} (+1.26) & \second{95.29} (+8.81) & \second{21.81} (-39.12) & \second{70.49} (+5.55) \\
                \rowcolor[HTML]{EFEFEF} 
                + UniEnt+ & \best{85.18} (+4.26) & \second{96.97} (+12.65) & \second{14.28} (-57.38) & \best{83.67} (+11.04) & 59.71(-0.92) & \best{94.23} (+5.59) & \best{26.87} (-23.31) & \best{58.19} (+0.95) & \best{72.45} (+1.67) & \best{95.60} (+9.12) & \best{20.58} (-40.35) & \best{70.93} (+6.00) \\ \midrule
                OSTTA~\cite{lee2023towards} & \best{84.44} & 72.74 & 77.02 & 65.17 & \best{60.03} & 75.37 & 82.75 & 51.35 & \best{72.24} & 74.06 & 79.89 & 58.26 \\
                \rowcolor[HTML]{EFEFEF} 
                + UniEnt & 82.46 (-1.98) & \second{96.20} (+23.46) & \second{16.37} (-60.65) & \second{80.51} (+15.34) & 58.69 (-1.34) & \second{94.84} (+19.47) & \second{22.95} (-59.80) & \second{57.28} (+5.93) & 70.58 (-1.66) & \second{95.52} (+21.47) & \second{19.66} (-60.23) & \second{68.90} (+10.64) \\
                \rowcolor[HTML]{EFEFEF} 
                + UniEnt+ & \second{84.30} (-0.14) & \best{97.38} (+24.64) & \best{11.56} (-65.46) & \best{82.91} (+17.74) & \second{58.93} (-1.10) & \best{95.42} (+20.05) & \best{20.59} (-62.16) & \best{57.69} (+6.34) & \second{71.62} (-0.62) & \best{96.40} (+22.35) & \best{16.08} (-63.81) & \best{70.30} (+12.04) \\ \bottomrule
            \end{tabular}%
        }
    }
    \\
    \subfloat[Results after 10 rounds of adaptation (\ie, long-term test-time adaptation).]{
        \resizebox{\linewidth}{!}{%
            \begin{tabular}{l|cccc|cccc|cccc}
                \toprule
                \multirow{2}{*}{Method} & \multicolumn{4}{c|}{CIFAR-10-C} & \multicolumn{4}{c|}{CIFAR-100-C} & \multicolumn{4}{c}{Average} \\ \cmidrule{2-13}
                 & Acc$\uparrow$ & AUROC$\uparrow$ & FPR@TPR95$\downarrow$ & OSCR$\uparrow$ & Acc$\uparrow$ & AUROC$\uparrow$ & FPR@TPR95$\downarrow$ & OSCR$\uparrow$ & Acc$\uparrow$ & AUROC$\uparrow$ & FPR@TPR95$\downarrow$ & OSCR$\uparrow$ \\ \midrule
                Source~\cite{zagoruyko2016wide} & 81.73 & 77.89 & 79.45 & 68.44 & 53.25 & 60.55 & 94.98 & 39.87 & 67.49 & 69.22 & 87.22 & 54.16 \\
                BN Adapt~\cite{nado2020evaluating} & 84.20 & 80.40 & 76.84 & 72.13 & 57.16 & 72.45 & 84.28 & 47.09 & 70.68 & 76.43 & 80.57 & 59.62 \\
                CoTTA~\cite{wang2022continual} & 35.90 & 47.27 & 97.52 & 19.95 & 13.34 & 48.34 & 91.61 & 8.19 & 24.62 & 47.81 & 94.57 & 14.07 \\ \midrule
                TENT~\cite{wang2021tent} & 32.61 & 60.86 & 93.24 & 20.86 & 37.49 & 53.73 & 95.07 & 25.02 & 35.05 & 57.30 & 94.16 & 22.94 \\
                \rowcolor[HTML]{EFEFEF} 
                + UniEnt & \second{84.07} (+51.46) & \best{88.53} (+27.67) & \best{51.48} (-41.76) & \best{77.87} (+57.01) & \best{57.93} (+20.44) & \second{90.62} (+36.89) & \second{46.18} (-48.89) & \best{55.67} (+30.65) & \second{71.00} (+35.95) & \best{89.58} (+32.28) & \best{48.83} (-45.33) & \best{66.77} (+43.83) \\
                \rowcolor[HTML]{EFEFEF} 
                + UniEnt+ & \best{84.17} (+51.56) & \second{88.21} (+27.35) & \second{52.57} (-40.67) & \second{77.75} (+56.89) & \second{57.92} (+20.43) & \best{90.63} (+36.90) & \best{45.10} (-49.97) & \second{55.59} (+30.57) & \best{71.05} (+36.00) & \second{89.42} (+32.13) & \second{48.84} (-45.32) & \second{66.67} (+43.73) \\ \midrule
                EATA~\cite{niu2022efficient} & 40.94 & 64.52 & 88.41 & 29.07 & 48.75 & 73.26 & 80.83 & 41.27 & 44.85 & 68.89 & 84.62 & 35.17 \\
                \rowcolor[HTML]{EFEFEF} 
                + UniEnt & \best{81.22} (+40.28) & \second{91.05} (+26.53) & \second{30.59} (-57.82) & \second{76.42} (+47.35) & \second{57.07} (+8.32) & \best{98.59} (+25.33) & \best{5.85} (-74.98) & \second{56.70} (+15.43) & \second{69.15} (+24.30) & \second{94.82} (+25.93) & \best{18.22} (-66.40) & \second{66.56} (+31.39) \\
                \rowcolor[HTML]{EFEFEF} 
                + UniEnt+ & \second{80.41} (+39.47) & \best{92.49} (+27.97) & \best{30.00} (-58.41) & \best{77.00} (+47.93) & \best{58.02} (+9.27) & \second{98.05} (+24.79) & \second{7.92} (-72.91) & \best{57.47} (+16.20) & \best{69.22} (+24.37) & \best{95.27} (+26.38) & \second{18.96} (-65.66) & \best{67.24} (+32.07) \\ \midrule
                OSTTA~\cite{lee2023towards} & \best{83.83} & 71.93 & 76.12 & 63.90 & \second{57.39} & 75.46 & 82.47 & 49.61 & \best{70.61} & 73.70 & 79.30 & 56.76 \\
                \rowcolor[HTML]{EFEFEF} 
                + UniEnt & 80.74 (-3.09) & \second{88.94} (+17.01) & \second{35.66} (-40.46) & \second{74.52} (+10.62) & 56.13 (-1.26) & \second{95.20} (+19.74) & \second{21.15} (-61.32) & \second{54.89} (+5.28) & 68.44 (-2.18) & \second{92.07} (+18.38) & \second{28.41} (-50.89) & \second{64.71} (+7.95) \\
                \rowcolor[HTML]{EFEFEF} 
                + UniEnt+ & \second{82.42} (-1.41) & \best{90.15} (+18.22) & \best{31.18} (-44.94) & \best{76.46} (+12.56) & \best{57.45} (+0.06) & \best{95.91} (+20.45) & \best{17.33} (-65.14) & \best{56.32} (+6.71) & \second{69.94} (-0.67) & \best{93.03} (+19.34) & \best{24.26} (-55.04) & \best{66.39} (+9.64) \\ \bottomrule
            \end{tabular}%
        }
    }
    \caption{Results of different methods on CIFAR benchmarks.}
    \label{tab:long-term}
\end{table*}

\paragraph{Effects of learning rate and batch size.}
We explore the impact of learning rate and batch size on our approaches in \Tref{tab:lr&bs}. A learning rate that is too large or too small can hurt performance, while a larger batch size results in better performance. Compared to TENT~\cite{wang2021tent} and EATA~\cite{niu2022efficient}, our methods are more robust to learning rate and batch size. Nonetheless, our methods share the same limitation as the baseline methods: they rely on a large batch size to estimate the distribution accurately. Moreover, we observe that OSTTA~\cite{lee2023towards} is less sensitive to learning rate and batch size.

\begin{table*}[t]
    \centering
    \subfloat[OSCR \wrt learning rate]{
        \resizebox{0.48\linewidth}{!}{%
            \begin{tabular}{l|cccc|c}
                \toprule
                \multirow{2}{*}{Method} & \multicolumn{4}{c|}{Learning rate} & \multirow{2}{*}{$\Delta$} \\ \cmidrule(lr){2-5}
                 & 0.005 & 0.001 & 0.0005 & 0.0001 &  \\ \midrule
                Source~\cite{zagoruyko2016wide} & 39.87 & 39.87 & 39.87 & 39.87 & 0.00 \\
                BN Adapt~\cite{nado2020evaluating} & 47.10 & 47.10 & 47.10 & 47.10 & 0.00 \\ \midrule
                TENT~\cite{wang2021tent} & 10.60 & 42.24 & 42.38 & 48.36 & 37.76 \\
                \rowcolor[HTML]{EFEFEF} 
                + UniEnt & \second{53.82} (+43.22) & \second{56.20} (+13.96) & \second{56.06} (+13.68) & \second{54.51} (+6.15) & 2.38 \\
                \rowcolor[HTML]{EFEFEF} 
                + UniEnt+ & \best{54.44} (+43.84) & \best{56.36} (+14.12) & \best{56.27} (+13.89) & \best{54.65} (+6.29) & 1.92 \\ \midrule
                EATA~\cite{niu2022efficient} & 40.96 & 57.00 & 56.91 & \second{53.60} & 16.04 \\
                \rowcolor[HTML]{EFEFEF} 
                + UniEnt & \best{49.36} (+8.40) & \second{57.76} (+0.76) & \second{57.10} (+0.19) & \best{53.63} (+0.03) & 8.40 \\
                \rowcolor[HTML]{EFEFEF} 
                + UniEnt+ & \second{49.05} (+8.09) & \best{58.07} (+1.07) & \best{57.39} (+0.48) & 53.40 (-0.20) & 9.02 \\ \midrule
                OSTTA~\cite{lee2023towards} & 49.43 & 51.35 & 51.98 & 52.37 & 2.94 \\
                \rowcolor[HTML]{EFEFEF} 
                + UniEnt & \second{51.41} (+1.98) & \second{56.93} (+5.58) & \second{57.22} (+5.24) & \second{55.58} (+3.21) & 5.81 \\
                \rowcolor[HTML]{EFEFEF} 
                + UniEnt+ & \best{53.39} (+3.96) & \best{57.69} (+6.34) & \best{57.68} (+5.70) & \best{56.06} (+3.69) & 4.30 \\ \bottomrule
            \end{tabular}%
        }
    }
    \hfill
    \subfloat[OSCR \wrt batch size]{
        \resizebox{0.48\linewidth}{!}{%
            \begin{tabular}{l|cccc|c}
                \toprule
                \multirow{2}{*}{Method} & \multicolumn{4}{c|}{Batch size} & \multirow{2}{*}{$\Delta$} \\ \cmidrule(lr){2-5}
                 & 64 & 32 & 16 & 8 &  \\ \midrule
                Source~\cite{zagoruyko2016wide} & 39.87 & 39.87 & 39.87 & 39.87 & 0.00 \\
                BN Adapt~\cite{nado2020evaluating} & 46.38 & 45.25 & 42.94 & 38.61 & 7.77 \\ \midrule
                TENT~\cite{wang2021tent} & 33.27 & 8.10 & 2.51 & 0.95 & 32.32 \\
                \rowcolor[HTML]{EFEFEF} 
                + UniEnt & \best{55.17} (+21.90) & \second{53.05} (+44.95) & \second{48.87} (+46.36) & \best{31.47} (+30.52) & 23.70 \\
                \rowcolor[HTML]{EFEFEF} 
                + UniEnt+ & \second{55.17} (+21.90) & \best{53.13} (+45.03) & \best{49.27} (+46.76) & \second{28.35} (+27.40) & 26.82 \\ \midrule
                EATA~\cite{niu2022efficient} & 53.09 & 47.78 & 40.57 & 31.57 & 21.52 \\
                \rowcolor[HTML]{EFEFEF} 
                + UniEnt & \best{57.08} (+3.99) & \best{54.52} (+6.74) & \best{50.71} (+10.14) & \best{43.89} (+12.32) & 13.19 \\
                \rowcolor[HTML]{EFEFEF} 
                + UniEnt+ & \second{56.79} (+3.70) & \second{54.29} (+6.51) & \second{50.49} (+9.92) & \second{43.17} (+11.60) & 13.62 \\ \midrule
                OSTTA~\cite{lee2023towards} & 50.35 & 48.82 & \second{46.07} & \second{39.75} & 10.60 \\
                \rowcolor[HTML]{EFEFEF} 
                + UniEnt & \second{54.54} (+4.19) & \second{50.49} (+1.67) & 44.97 (-1.10) & 36.72 (-3.03) & 17.82 \\
                \rowcolor[HTML]{EFEFEF} 
                + UniEnt+ & \best{55.76} (+5.41) & \best{52.66} (+3.84) & \best{47.94} (+1.87) & \best{41.45} (+1.70) & 14.31 \\ \bottomrule
            \end{tabular}%
        }
    }
    \caption{OSCR of different methods on CIFAR-100-C with diverse learning rates and batch sizes. $\Delta$ is the difference between the largest and smallest values. Smaller $\Delta$ values represent better robustness.}
    \label{tab:lr&bs}
\end{table*}

\paragraph{Effects of OOD score.}
We use the energy score~\cite{liu2020energy} to measure the model's detection performance on csOOD data. From \Tref{tab:score}, we can make two observations. First, our methods consistently improve the performance using different OOD scores. Second, compared with MSP~\cite{hendrycks2017baseline}, using Max Logit~\cite{hendrycks2022scaling} and Energy yields better detection performance.

\begin{table}[t]
    \centering
    \resizebox{\linewidth}{!}{%
        \begin{tabular}{l|ccc|c}
            \toprule
            \multirow{2}{*}{Method} & \multicolumn{3}{c|}{OOD score} & \multirow{2}{*}{$\Delta$} \\ \cmidrule(lr){2-4}
             & MSP~\cite{hendrycks2017baseline} & Max Logit~\cite{hendrycks2022scaling} & Energy~\cite{liu2020energy} &  \\ \midrule
            Source~\cite{zagoruyko2016wide} & 39.65 & 40.24 & 39.87 & 0.59 \\
            BN Adapt~\cite{nado2020evaluating} & 48.75 & 48.04 & 47.10 & 1.65 \\
            CoTTA~\cite{wang2022continual} & 49.44 & 49.73 & 48.99 & 0.74 \\ \midrule
            TENT~\cite{wang2021tent} & 36.86 & 41.79 & 42.24 & 5.38 \\
            \rowcolor[HTML]{EFEFEF} 
            + UniEnt & \best{55.42} (+18.56) & \second{56.20} (+14.41) & \second{56.26} (+14.02) & 0.84 \\
            \rowcolor[HTML]{EFEFEF} 
            + UniEnt+ & \second{55.24} (+18.38) & \best{56.31} (+14.52) & \best{56.36} (+14.12) & 1.12 \\ \midrule
            EATA~\cite{niu2022efficient} & 55.20 & 57.52 & 57.55 & 2.35 \\
            \rowcolor[HTML]{EFEFEF} 
            + UniEnt & \second{56.94} (+1.74) & \second{57.88} (+0.36) & \second{57.87} (+0.32) & 0.94 \\
            \rowcolor[HTML]{EFEFEF} 
            + UniEnt+ & \best{57.37} (+2.17) & \best{58.33} (+0.81) & \best{58.33} (+0.78) & 0.96 \\ \midrule
            OSTTA~\cite{lee2023towards} & 49.14 & 51.42 & 51.35 & 2.28 \\
            \rowcolor[HTML]{EFEFEF} 
            + UniEnt & \second{56.52} (+7.38) & \second{57.23} (+5.81) & \second{57.25} (+5.90) & 0.73 \\
            \rowcolor[HTML]{EFEFEF} 
            + UniEnt+ & \best{57.12} (+7.98) & \best{57.69} (+6.27) & \best{57.69} (+6.34) & 0.57 \\ \bottomrule
        \end{tabular}%
    }
    \caption{OSCR of different methods on CIFAR-100-C using diverse OOD scores.}
    \label{tab:score}
\end{table}

\end{document}